%% file: main.tex
\RequirePackage[svgnames,table]{xcolor}
\documentclass[11pt,letterpaper,logo]{yalearxiv}

\input{preamble}
\input{university_logos} 

\title{Verifying Visual Claims Step by Step\\
for On-Policy Distillation}

\runningtitle{VV-OPD: Verifying Visual Claims Step by Step for On-Policy Distillation}

\author{%
{\fontsize{11}{13}\selectfont\bfseries
Peizhen Li$^{1}$, Zehong Cao$^{1}$, Yugu Li$^{1}$, Jianglin Qiao$^{2}$}\\
{\fontsize{11}{13}\selectfont\bfseries
Siyi Hu$^{3}$, Xiao-Ming Wu$^{4}$, Yang Zhang$^{5}$}\vspace{0.5em}\\
{\small
$^{1}$Adelaide University \quad
$^{2}$The University of Sydney \quad
$^{3}$Curtin University}\vspace{0.15em}\\
{\small
$^{4}$Nanyang Technological University \quad
$^{5}$University of North Texas}%
}

\hypersetup{
  colorlinks=true,
  linkcolor=blue!50!black,
  citecolor=blue!50!black,
  urlcolor=blue!50!black,
  pdftitle={Verifying Visual Claims Step by Step for On-Policy Distillation},
  pdfauthor={Peizhen Li, Zehong Cao, Yugu Li, Jianglin Qiao, Siyi Hu, Xiao-Ming Wu, Yang Zhang}
}

\begin{document}

\begin{abstract}
{\centering\section*{Abstract}}

On-policy distillation (OPD) for vision–language reasoning can transfer a teacher’s visual errors to the student. When the teacher misidentifies an object attribute, misinterprets a spatial relation, or misreads a scene, the resulting erroneous claims can become supervision targets for the student’s intermediate reasoning steps. Existing vision-based OPD methods reweight or redirect the distillation signal toward image content, leaving individual visual claims in the teacher’s reasoning unverified. We propose Visual Verification for On-Policy Distillation (VV-OPD), which checks whether the image supports the visual claims made at each step of the teacher’s reasoning. VV-OPD uses the verification results to construct contrastive pairs that train the student policy to favor reasoning steps grounded in visual evidence. Across eight benchmarks, VV-OPD improves over OPD by 6.86 points on mathematical reasoning and 4.37 points on general multimodal reasoning, while reducing step-level visual evidence errors by 3.34 and 2.34 points, respectively. Our findings suggest that step-level visual verification can make distillation more reliable by reducing the propagation of teacher errors while improving the student’s final-answer accuracy. Code is available at \url{https://github.com/lipzh5/VV-OPD}.

\end{abstract}

\maketitle

\section{Introduction}
\label{sec:intro}

\begin{figure}[h]
  \centering
  \includegraphics[width=0.70\linewidth]{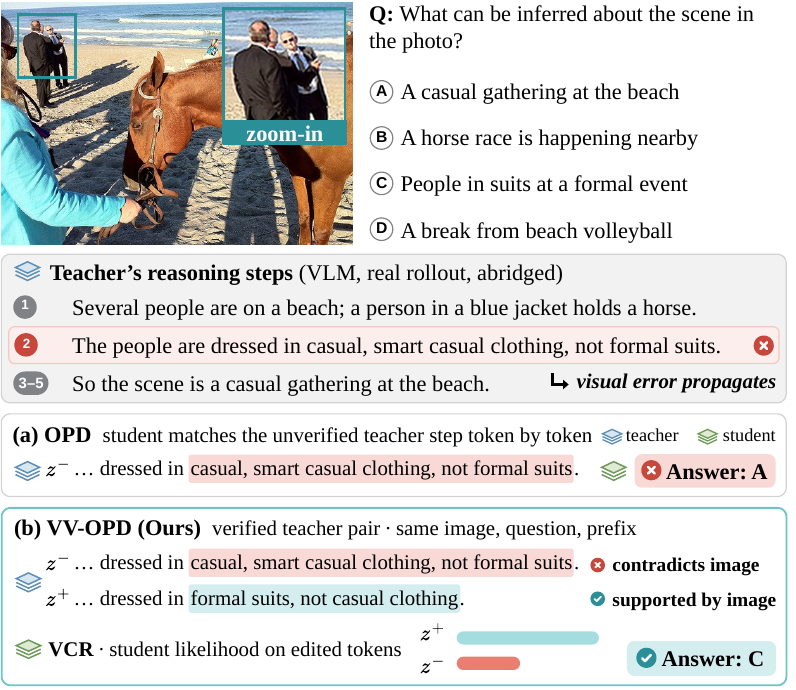}
  
\caption{\textbf{From teacher visual errors to corrective supervision.}
\textbf{(a)} Standard OPD reinforces the teacher’s misreading of formal suits as casual clothing, leading the student to select A.
\textbf{(b)} VV-OPD pairs the erroneous step $z^-$ with a verified correction $z^+$. Visual Consistency Ranking (VCR) uses the edited tokens to rank $z^+$ above $z^-$, and the student selects C.}

  \label{fig:motivation}
\end{figure}

On-policy distillation (OPD)~\cite{agarwal2024policy} has emerged to improve reasoning in vision–language models (VLMs) by providing teacher supervision along trajectories sampled from the student’s current policy. The student generates a response to an image–question pair, and the teacher provides a next-token distribution at each prefix of that response. The student thus learns from the teacher in the reasoning contexts it encounters during generation. VOLD~\cite{bousselham2026vold} is a representative framework that uses OPD to transfer reasoning capabilities from a text-only teacher to a vision–language student. In vision–language reasoning, the student must interpret visual evidence and use it to draw conclusions over multiple steps. For example, determining whether a parked car blocks a bike lane requires identifying the lane boundaries, locating the car, and checking whether it extends into the lane.

When OPD uses a vision–language teacher, the reliability of its visual claims matters. Even strong VLMs can misidentify object attributes, misinterpret spatial relations, or describe scenes in ways that conflict with the image~\cite{li2023evaluating,leng2024mitigating,guan2024hallusionbench,rohrbach2018object}. In the bike lane example, if the teacher misidentifies the lane boundary, its next-token predictions may reinforce that error in the student’s reasoning and lead to an incorrect conclusion about whether the car blocks the lane. Recent vision-based OPD methods, such as Vision-OPD and VA-OPD, strengthen visual supervision by giving the teacher access to evidence-centered image crops~\cite{yuan2026vision}, increasing the weight of visually informative tokens~\cite{liu2026visual}, or steering gradients toward visual grounding~\cite{yoon2026decomposed}. These approaches increase the influence of visual information on the distillation signal, leaving individual teacher claims unverified. This raises a question at each reasoning step: \textit{does the image support the teacher’s visual claim?}

Inspired by process reward models that evaluate intermediate reasoning steps~\cite{min2026cvpr,lightman2024let,wang2024math}, we propose \textbf{Visual Verification for On-Policy Distillation (VV-OPD)}, which verifies teacher visual claims at the step level and uses the results to construct verified contrastive pairs. As shown in Fig.~\ref{fig:motivation}, VV-OPD pairs a potentially erroneous step from a recorded teacher trajectory with a local correction under the same image, question, and reasoning prefix. An independent verifier retains the pair only if the original claim contradicts the image, the corrected claim is supported by the image, and the rest of the step is preserved. The student learns from these pairs through a reference-free \emph{visual consistency ranking} (VCR) loss, which uses the student's likelihood on the edited tokens to rank the corrected step above the original teacher step. This focuses direct supervision on the disputed visual claim. VCR is \emph{reward-gated} and applies only to questions for which at least one student rollout yields an incorrect answer. VV-OPD thus combines standard OPD supervision on student-generated rollouts with targeted corrections derived from verified teacher errors.

\noindent\textbf{Our contributions are threefold.}
\begin{itemize}
  \item We characterize \emph{visual reasoning error transfer} in OPD: token-level teacher matching can turn a visual claim contradicted by the image into a supervision target, encouraging the student to reproduce the claim and use it as a premise for subsequent reasoning.
  \item We propose \textit{VV-OPD}, which pairs visually inconsistent teacher steps with verified, image-grounded corrections under the same reasoning context. A reward-gated VCR loss trains the student to prefer the corrected steps by comparing likelihoods on the edited tokens.
  \item Across eight benchmarks, VV-OPD outperforms VA-OPD and Vision-OPD and improves by 6.86 points on mathematical reasoning and 4.37 points on multimodal reasoning relative to OPD, while reducing step-level visual evidence errors by 3.34 and 2.34 points, respectively.
\end{itemize}

\section{Related Work}
\label{sec:related-work}

\noindent\textbf{OPD for VLM reasoning.}
VOLD~\cite{bousselham2026vold} applies OPD~\cite{agarwal2024policy,gu2024minillm,ma2026mopd}
to multimodal reasoning with a text-only teacher, which cannot check the
student's claims against the image. Later methods strengthen visual
supervision: Vision-OPD~\cite{yuan2026vision} conditions the teacher on
evidence-centered crops, VA-OPD~\cite{liu2026visual} emphasizes tokens whose
teacher scores depend on visual details, and decomposed
OPD~\cite{yoon2026decomposed} steers optimization toward visual grounding.
These methods improve how visual information is transferred, but all of them
treat the teacher's claims as correct targets. A teacher that attends to the
image can still misidentify an object attribute or spatial
relation~\cite{guan2024hallusionbench}, and OPD then distills the error.
Inspired by step-level verification~\cite{lightman2024let,wang2024math},
VV-OPD checks individual teacher claims against the image and trains the
student to rank verified corrections above the erroneous steps.

\noindent\textbf{Granularity of visual supervision in VLM reasoning.}
Prior work evaluates object hallucinations~\cite{rohrbach2018object,li2023evaluating}
and fine-grained visual facts~\cite{jing2024faithscore}, or mitigates hallucinations
through contrastive decoding~\cite{leng2024mitigating}, data
cleaning~\cite{yu2024hallucidoctor}, contrastive
learning~\cite{jiang2024hallucination}, and preference
optimization~\cite{wang2024mdpo}.
\cite{uppaal2026journey} show that correct answers can coexist
with visually unfaithful intermediate reasoning, while
\cite{min2026cvpr} identify perceptual errors through
perception-centric process supervision. However, these step-level signals
are primarily used for evaluation, reward modeling, or inference-time
correction rather than correcting teacher supervision during distillation.
RLHF-V~\cite{yu2024rlhfv} corrects hallucinated response segments, whereas
Step-DPO~\cite{lai2024step} learns preferences between whole reasoning steps.
VV-OPD instead contrasts actual teacher steps with independently verified
local corrections under the same reasoning context, targeting only the
disputed visual claim during distillation.

\section{Method}
\label{sec:method}

\begin{figure*}[t]
  \centering
  \includegraphics[width=\textwidth]{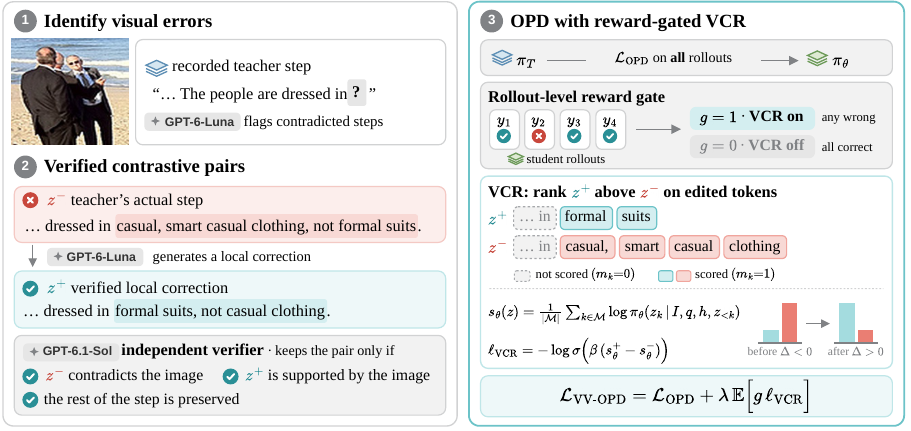}
  \caption{\textbf{Overview of VV-OPD.} From recorded teacher trajectories, GPT-6-Luna (1)~flags teacher steps whose visual claims contradict the image and (2)~generates a local correction $z^+$ for each flagged step $z^-$. GPT-6.1-Sol independently verifies each pair and keeps it only if $z^-$ contradicts the image, $z^+$ is supported by the image, and the rest of the step is preserved. (3)~a rollout-level reward gate sets $g{=}1$ when at least one student rollout for the question is incorrect. For gated questions, VCR ranks $z^+$ above $z^-$ using
only the edited tokens ($m_k{=}1$), encouraging a
positive score margin $\Delta$.
The reward-gated VCR loss is combined with OPD
on student rollouts.}
  \label{fig:vipod_framework}
\end{figure*}

VV-OPD reduces the transfer of teacher claims that contradict the image during distillation, as shown in Figure~\ref{fig:vipod_framework}. We first explain how matching the teacher's token distributions can reinforce visual errors in the student's reasoning (Sec.~\ref{sec:vis-error-transfer}). We then pair erroneous steps from recorded teacher trajectories with verified, image-grounded corrections (Sec.~\ref{sec:vipod-preferences}). Finally, we use these step-level contrastive pairs to train the student with reward-gated VCR alongside the OPD objective (Sec.~\ref{sec:vipod-objective}).

\subsection{Visual Error Transfer in OPD}
\label{sec:vis-error-transfer}

Given a vision--language input $x=(I,q)$, a trainable student $\pi_\theta$ samples a token trajectory $y=(y_1,\ldots,y_{|y|})\sim\pi_\theta(\cdot\mid x)$. At each student-generated prefix $y_{<t}$, a frozen teacher VLM $\pi_T$ provides a next-token distribution. Standard OPD updates the student by minimizing
\begin{equation}
\label{eq:vipod-opd}
\mathcal{L}_{\mathrm{OPD}}(\theta)
=
\mathbb{E}_{x\sim\mathcal{D},\,y\sim\pi_\theta(\cdot\mid x)}
\left[
\frac{1}{|y|}\sum_{t=1}^{|y|}
\mathrm{KL}_K(p_T^t\,\|\,p_\theta^t)
\right],
\end{equation}
where $p_T^t=\pi_T(\cdot\mid x,y_{<t})$ and $p_\theta^t=\pi_\theta(\cdot\mid x,y_{<t})$ are the teacher and student next-token distributions, and $\mathrm{KL}_K$ denotes the top-$K$ approximation to forward KL.

\noindent\textbf{How visual errors are transferred.}
As shown in Fig.~\ref{fig:vipod_framework}, at the prefix ``People are dressed'', a teacher that mistakes formal suits for casual clothing may favor a continuation that describes the clothing as casual. The distillation objective uses the teacher's probabilities as targets, so this visual misinterpretation can influence the student's next-token predictions. For full-vocabulary forward KL, the gradient with respect to the student's pre-softmax logit $\ell_v^t$ for token $v$ is
\begin{equation}
\label{eq:opd-grad}
\frac{\partial\,\mathrm{KL}(p_T^t\,\|\,p_\theta^t)}
{\partial \ell_v^t}
=
p_\theta^t(v)-p_T^t(v).
\end{equation}
Thus, if the teacher assigns more probability than the student to a token that favors an erroneous continuation, gradient descent increases that token's logit. At a fixed prefix, the unrestricted full-vocabulary optimum matches the teacher's probability mass on any set $\mathcal{V}_I^-$ of next tokens that favor a visually contradicted continuation:
\begin{equation}
\label{eq:opd-mass}
p_\theta^{t,*}(\mathcal{V}_I^-)=p_T^t(\mathcal{V}_I^-).
\end{equation}
These identities describe full-vocabulary KL. The top-$K$ objective approximates teacher matching on a selected token support. In both cases, the teacher distribution supplies the target, so a continuation that contradicts the image can still receive a positive training signal. We examine the downstream effects of this error-transfer mechanism in Sec.~\ref{sec:training_dynamics}.

\noindent\textbf{Identifying teacher errors for correction.}
To obtain concrete examples of this failure mode, we examine recorded teacher trajectories separately from the student rollouts used for OPD. Given a step-structured teacher trajectory $\tau_T=(u_1,\ldots,u_m)$ for $(I,q)$, we identify steps that contain a visual claim contradicted by the image, such as an incorrect object attribute or spatial relation. We restrict this analysis to errors that involve a visual claim. For each identified step $u_j$, we set $z_j^-=u_j$ and retain its preceding reasoning context $h_j=(u_1,\ldots,u_{j-1})$.


\subsection{Verified Contrastive Pair Construction}
\label{sec:vipod-preferences}

\noindent\textbf{Local correction.}
For each rejected step $z_j^-$, we construct an image-grounded correction $z_j^+$ under the same context $(I,q,h_j)$. The correction focuses on the contradicted visual claim while preserving the surrounding reasoning and the original prefix $h_j$. Rather than rewriting the entire reasoning trajectory, we modify only the relevant local content. Such corrections are not necessarily single-token replacements; for example, correcting an object attribute or specifying the subject of a visual claim may require editing several tokens.

\noindent\textbf{Independent verification.}
A separate verifier evaluates each candidate pair against the input image and its reasoning context. A pair is retained only if (i) $z_j^-$ contains a visual claim contradicted by the image, (ii) $z_j^+$ is supported by the visual evidence, and (iii) the correction preserves the local reasoning context without unnecessary modifications. Pairs failing any of these conditions are discarded. As illustrated in Fig.~\ref{fig:vipod_framework}, the retained pairs form
\begin{equation}
\label{eq:vipod-pairs}
\mathcal{P}=\left\{(I_i,q_i,h_i,z_i^+,z_i^-)\right\}_{i=1}^{N}.
\end{equation}
Each pair therefore provides a localized contrast between an actual teacher error and its verified, image-grounded alternative, rather than between unrelated reasoning trajectories.

\noindent\textbf{Models for pair construction.}
We use GPT-6-Luna for identifying visually contradicted teacher steps and generating local corrections, and GPT-6.1-Sol for independent verification. The verified pairs are constructed offline from recorded teacher trajectories, separately from the student-generated rollouts used during distillation. 


\subsection{Reward-Gated Visual Consistency Ranking}
\label{sec:vipod-objective}

\noindent\textbf{Edit-token masking and step-level scoring.}
For a fixed pair $p=(I,q,h,z^+,z^-)\in\mathcal{P}$, let
$z\in\{z^+,z^-\}$ denote either candidate, represented as a token
sequence $z=(z_1,\ldots,z_{|z|})$. We denote its edited token positions,
identified relative to the other candidate in $p$, by
$\mathcal{M}(z)$. For insertions or deletions, aligned boundary tokens
are included so that both candidates have nonempty scored positions.
We define the masked, length-normalized score
\begin{equation}
\label{eq:VV-OPD-score}
s_\theta(z)
=\frac{1}{|\mathcal{M}(z)|}
\sum_{k\in\mathcal{M}(z)}
\log\pi_\theta(z_k\mid I,q,h,z_{<k}).
\end{equation}
Here the pair $p$, and hence $(I,q,h)$ and $\mathcal{M}(z)$, is fixed
when evaluating $s_\theta(z)$. Unedited tokens remain in the
conditioning sequence but do not contribute to the score.
Normalizing by $|\mathcal{M}(z)|$ prevents differences in edit length
from directly scaling the score.

\noindent\textbf{Visual consistency ranking (VCR).}
For each pair $p$, we define
$\Delta_\theta(p)=s_\theta(z^+)-s_\theta(z^-)$ and minimize
\begin{equation}
\label{eq:vipod-vcr}
\ell_{\mathrm{VCR}}(\theta;p)
=-\log\sigma\!\left(\beta\Delta_\theta(p)\right),
\end{equation}
where $\beta>0$ controls the sharpness of the ranking loss.
This objective favors the image-grounded correction $z^+$ over the
visually contradicted teacher step $z^-$. Unlike offline preference optimization~\cite{rafailov2023direct}, VCR is
applied on-policy alongside distillation; like SimPO~\cite{meng2024simpo}, it
is reference-free and requires neither a reference model nor teacher
likelihoods for the candidate steps.

\noindent\textbf{Reward gating.}
For each training question $x=(I,q)$, the student samples a group
$\mathcal{Y}_x$ of on-policy rollouts. Let $r(x,y)\in\{0,1\}$ indicate
final-answer correctness and $\mathcal{P}_x\subseteq\mathcal{P}$
denote the verified pairs associated with $x$. We activate VCR only
when a pair is available and at least one rollout is incorrect:
\begin{equation}
\label{eq:vipod-gate}
g(x,\mathcal{Y}_x)
=\mathbb{I}[\mathcal{P}_x\ne\varnothing]\,
\mathbb{I}\!\left[\exists y\in\mathcal{Y}_x:
                   r(x,y)=0\right].
\end{equation}
The gate $g(x,\mathcal{Y}_x)\in\{0,1\}$ controls only the auxiliary
ranking term. For each active question ($g=1$), we sample one
$p_x\in\mathcal{P}_x$ per update; for inactive questions ($g=0$), we skip the pair and set the VCR contribution to zero.

\noindent\textbf{VV-OPD objective.}
The student is trained by combining on-policy KL distillation with
reward-gated VCR:
\begin{equation}
\label{eq:vipod-total}
\begin{aligned}
\mathcal{L}_{\mathrm{VV-OPD}}(\theta)
={}&\mathbb{E}_{x,\mathcal{Y}_x}\Bigg[
\frac{1}{|\mathcal{Y}_x|}
\sum_{y\in\mathcal{Y}_x}
\frac{1}{|y|}
\sum_{t=1}^{|y|}
\mathrm{KL}_K(p_T^t\|p_\theta^t)\\
&\qquad+\lambda\,g(x,\mathcal{Y}_x)
\,\ell_{\mathrm{VCR}}(\theta;p_x)
\Bigg].
\end{aligned}
\end{equation}
Here $x\sim\mathcal{D}$, and $\mathcal{Y}_x$ consists of student
rollouts sampled from $\pi_\theta(\cdot\mid x)$. Within the inner
sum, $p_T^t$ and $p_\theta^t$ are evaluated on the prefix of the
particular $y$. The expectation also includes the random choice of
$p_x$ for active questions. The coefficient $\lambda\ge0$ weights
VCR; both losses update the same student parameters $\theta$.
Once the pairs are constructed offline, VV-OPD requires no additional
teacher evaluations or student rollouts beyond OPD; its online
overhead is scoring one verified pair per active question.

\section{Experiments and Results}
\label{sec:exp}
\subsection{Experimental Setup}
\label{sec:setup}

\begin{figure}[t]
  \centering
  \includegraphics[width=.6\linewidth]{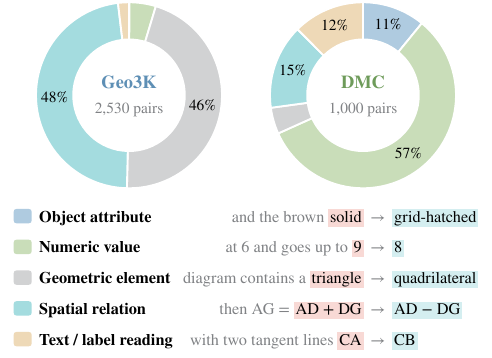}
  \caption{\textbf{Distribution of local correction types in verified
  contrastive pairs.} Donut charts show the proportions of correction types in
  Geo3K (left) and DMC (right), with each pair assigned one primary type.
  Representative examples below illustrate each type, highlighting the edited
  spans in the erroneous teacher step $z^-$ (red) and its verified correction
  $z^+$ (teal), while unchanged preceding text is shown in grey.}
  \label{fig:correction_types}
\end{figure}

\begin{table}[t]
\centering
\caption{\textbf{Main results on eight benchmarks.}
The primary setting uses Qwen3-VL-8B$\rightarrow$2B.
Scores are reported as Avg@4 (\%), with macro-averages
across four benchmarks per group.
Green labels denote absolute gains over Base 2B.}
\label{tab:main}

\footnotesize
\setlength{\tabcolsep}{2pt}
\renewcommand{\arraystretch}{1.08}

\resizebox{\linewidth}{!}{%
\begin{tabular}{@{}l*{10}{c}@{}}
\toprule
\multirow{2}{*}{\textbf{Method}}
& \multicolumn{5}{c}{
  \cellcolor{mathbg}\textbf{Math Reasoning}}
& \multicolumn{5}{c}{
  \cellcolor{visbg}\textbf{General Multimodal}} \\
\cmidrule(lr){2-6}
\cmidrule(l){7-11}

& MathVerse & WeMath & MV-MATH & MathVision
& \textbf{Avg}
& LogicVista & MMMU & MMMU-Pro & MMStar
& \textbf{Avg} \\
\midrule

Base 2B
& 33.37 & 55.09 & 17.97 & 18.28 & 31.18
& 40.21 & 42.47 & 33.71 & 59.23 & 43.91 \\

SFT
& 35.72 & 57.04 & 19.89 & 16.69 & 32.34
& 40.77 & 43.31 & 33.21 & 58.48 & 43.94 \\

GRPO
& 37.35 & 62.04 & 19.71 & 19.84 & 34.74
& 42.06 & 44.22 & 34.00 & 60.53 & 45.20 \\

OPD
& 34.59 & 58.94 & 19.79 & 21.16 & 33.62
& 43.96 & 41.67 & 34.03 & 58.72 & 44.60 \\

VOLD
& 27.78 & 54.02 & 10.65 & 12.29 & 26.19
& 32.38 & 45.31 & 31.36 & 54.22 & 40.82 \\

Vision-OPD
& 31.14 & 56.01 & 14.50 & 23.06 & 31.18
& 41.95 & 45.03 & 34.18 & 55.00 & 44.04 \\

VA-OPD
& 37.35 & 63.15 & \textbf{20.78} & 26.25 & 36.88
& 42.11 & 47.50 & 36.04 & 60.33 & 46.50 \\

\midrule

\textbf{VV-OPD (ours)}
& \textbf{42.96} & \textbf{69.97} & 20.31
& \textbf{28.67}
& \textbf{40.48}\,\scoreup{9.30}
& \textbf{47.09} & \textbf{48.50}
& \textbf{38.14} & \textbf{62.17}
& \textbf{48.97}\,\scoreup{5.06} \\

\bottomrule
\end{tabular}%
}
\end{table}

\noindent\textbf{Models and training data.}
Following VA-OPD~\cite{liu2026visual}, our primary setting uses
Qwen3-VL-8B-Instruct as the teacher and Qwen3-VL-2B-Instruct as the student~\cite{bai2025qwen3},
and we additionally evaluate 4B and 32B teachers with the same 2B student. The
student is trained on Geo3K~\cite{lu2021inter} with 2,530 verified step-level
contrastive pairs, constructed offline from the teacher's trajectories.
For cross-domain validation, we construct the DVQA--M3CoT Collection (DMC) from DVQA~\cite{kafle2018dvqa} and M3CoT~\cite{chen2024m3cot}, yielding 1,000 verified contrastive pairs. DMC is held out from training and used to evaluate the transfer of learned visual preferences and the recurrence of teacher visual reasoning errors.
To characterize the pairs, we use GPT-6.1-Terra to assign each pair one
primary correction type based on the image and the edited spans between $z^-$
and $z^+$, followed by human review of the classifications. As shown in Fig.~\ref{fig:correction_types}, the two sources cover
complementary visual reasoning errors: Geo3K corrections mostly concern spatial relations
(48\%) and geometric elements (46\%), whereas DMC corrections are dominated by
numeric values (57\%), followed by spatial relations, text and label reading,
and object attributes. The training questions serve as
prompts for the on-policy student rollouts used by OPD, while the offline pairs
provide step-level corrective supervision for VCR.

\noindent\textbf{Baselines and benchmarks.}
We compare against SFT~\cite{ouyang2022training,hu2021lora}, GRPO~\cite{shao2024deepseekmath},
standard OPD~\cite{agarwal2024policy}, and three distillation methods designed
for VLMs: VOLD~\cite{bousselham2026vold}, Vision-OPD~\cite{yuan2026vision}, and
VA-OPD~\cite{liu2026visual}. Since VV-OPD targets visually contradicted claims
in multi-step reasoning, we select benchmarks that require reasoning over visual
evidence rather than relying on textual information alone.
For mathematical reasoning, we use MathVerse~\cite{zhang2024mathverse}, WeMath~\cite{qiao2025we}, MV-MATH~\cite{wang2025mv}, and 
MathVision~\cite{wang2024measuring}. For general multimodal reasoning, we use
LogicVista~\cite{xiao2024logicvista},
MMMU~\cite{yue2024mmmu} and its harder variant MMMU-Pro~\cite{yue2025mmmu}, and
MMStar~\cite{chen2024we}.

\noindent\textbf{Evaluation metrics.}
We sample four responses per question at a decoding temperature of 1.0 and
report their mean final-answer accuracy (Avg@4) under benchmark-specific evaluation protocols, including GPT-based answer judging where applicable. Final-answer accuracy alone
cannot tell whether an answer rests on a correct reading of the
image~\cite{uppaal2026journey}, so we also measure visual reasoning errors at the step
level. For a benchmark, let $\mathcal{Z}$ denote the reasoning steps of all
sampled responses, $\mathcal{Z}_{\mathrm{perc}} \subseteq \mathcal{Z}$ the
perception steps that make claims about the image, and
$\mathcal{Z}_{\mathrm{err}} \subseteq \mathcal{Z}_{\mathrm{perc}}$ the steps
with at least one visual claim that contradicts the image. We report two
step-level visual reasoning error rates,
\begin{equation}
\verchain = \frac{|\mathcal{Z}_{\mathrm{err}}|}{|\mathcal{Z}|},
\verperc = \frac{|\mathcal{Z}_{\mathrm{err}}|}{|\mathcal{Z}_{\mathrm{perc}}|}.
\label{eq:ver}
\end{equation}
$\verchain$ measures how often a step of the reasoning chain contradicts the
image. Because a model can lower it simply by making fewer visual claims, we
also report $\verperc$, the error rate among the steps that do describe the
image, which corresponds to the unfaithful perception rate
of~\cite{uppaal2026journey}.

\noindent\textbf{Implementation details.}
We implement VV-OPD on veRL~\cite{sheng2025hybridflow} and train for 5 epochs
with AdamW, a constant learning rate of $10^{-6}$, a batch size of 16, and
$|\mathcal{Y}_x|=4$ student rollouts per question. We set $\lambda=0.001$ and
$\beta=2$. Further training details, as well as the models, prompts, and full
protocol used for the step-level evaluation, are provided in the Supplementary
Material.


\subsection{Main Results}
\label{sec:main-results}

\noindent\textbf{Overall performance.}
Tab.~\ref{tab:main} reports Avg@4 across eight benchmarks. VV-OPD consistently outperforms standard OPD, improving the Math Reasoning average from 33.62 to 40.48 (+6.86 points) and the General Multimodal average from 44.60 to 48.97 (+4.37 points). Relative to the vanilla 2B student, the gains reach 9.30 and 5.06 points, respectively. VV-OPD achieves the highest score on seven of the eight benchmarks. The largest improvements over OPD occur on WeMath (+11.03), MathVerse (+8.37), and MathVision (+7.51), where correctly interpreting visual elements is essential for subsequent mathematical reasoning. These results suggest that correcting local visual misinterpretations can benefit the entire reasoning process, as an incorrect visual premise may otherwise propagate through subsequent, mathematically valid steps.

\noindent\textbf{Comparison with existing training strategies.}
VV-OPD exceeds GRPO by 5.74 and 3.77 points on the two group averages, despite using answer correctness only to control when visual correction is applied, rather than as a policy-gradient signal. This distinction is important: an outcome reward evaluates whether the final answer is correct, but does not explicitly identify which intermediate visual claim is unsupported by the image. VV-OPD instead provides direct supervision on such claims through verified step-level contrastive pairs. Compared with VA-OPD, VV-OPD further improves the two averages by 3.60 and 2.47 points. In particular, the gains on MathVerse (+5.61) and WeMath (+6.82) suggest that emphasizing visual information alone may be insufficient when the teacher's interpretation of that information is incorrect. By explicitly contrasting image-grounded and contradicted reasoning steps, VV-OPD targets the correctness of visual supervision rather than its relative importance.

\noindent\textbf{Generalization beyond geometric reasoning.}
Although VV-OPD is trained on Geo3K, its improvements extend to general multimodal benchmarks. On MMMU and MMStar, standard OPD slightly underperforms the vanilla student (41.67 vs.\ 42.47 and 58.72 vs.\ 59.23), whereas VV-OPD achieves 48.50 and 62.17, respectively. Similar improvements appear on LogicVista (+3.13 over OPD) and MMMU-Pro (+4.11). This pattern suggests that the learned preference for visually supported reasoning is not restricted to the geometric concepts present in the training data. Instead, it may transfer to broader tasks requiring accurate interpretation of objects, diagrams, and visual relationships. The improvements are particularly encouraging given that the additional contrastive supervision is constructed solely from teacher errors on Geo3K.

\noindent\textbf{Performance on multi-image reasoning.}
The improvement is less pronounced on MV-MATH, where VV-OPD slightly trails VA-OPD (20.31 vs.\ 20.78), despite outperforming standard OPD (19.79). 
Unlike the predominantly single-image Geo3K training examples, MV-MATH often requires integrating information across multiple visual inputs, a challenge also addressed by multi-image preference optimization~\cite{li2025zooming}. Our verified pairs primarily correct localized visual claims under a shared reasoning context, which may provide less direct supervision for cross-image correspondence and information integration. This could explain the smaller gain on MV-MATH, although further experiments with multi-image contrastive pairs would be needed to verify this hypothesis.

\subsection{Visual Consistency}
\label{sec:visual-consistency}


\begin{table}[t]
\centering
\caption{\textbf{Step-level visual reasoning error rates (VER).}
Macro-averaged across benchmarks (\%; lower is better).
Chain and perc.\ denote all and perception steps.
Green labels show reductions over Base 2B.}
\label{tab:visual_consistency}

\footnotesize
\setlength{\tabcolsep}{2pt}
\renewcommand{\arraystretch}{1.06}

\newcommand{\verfullcell}[2]{%
  \mbox{%
    \textbf{#1}\hspace{2pt}%
    {\setlength{\fboxsep}{1pt}%
     \colorbox{vvteal!25}{%
       \textcolor{green!40!black}{\scriptsize$\downarrow$#2}%
     }}%
  }%
}

\begin{tabular*}{0.60\linewidth}{
@{\extracolsep{\fill}}lcccc@{}}
\toprule

\multirow{2}{*}{\textbf{Method}}
& \multicolumn{2}{c}{
  \cellcolor{mathbg}\textbf{Math Reasoning}}
& \multicolumn{2}{c}{
  \cellcolor{visbg}\textbf{General Multimodal}} \\
\cmidrule(lr){2-3}
\cmidrule(l){4-5}

& chain $\downarrow$
& perc.\ $\downarrow$
& chain $\downarrow$
& perc.\ $\downarrow$ \\
\midrule

Base 2B
& 23.82 & 48.13 & 20.03 & 40.64 \\

OPD
& 19.07 & 45.66 & 18.83 & 42.01 \\

SFT
& 28.09 & 55.39 & 26.97 & 46.32 \\

GRPO
& 18.24 & 45.54 & 17.72 & 47.45 \\

VOLD
& 30.93 & 61.78 & 29.04 & 54.45 \\

Vision-OPD
& 26.99 & 54.20 & 26.96 & 48.05 \\

VA-OPD
& 22.88 & 48.19 & 22.73 & 45.42 \\

\midrule

\textbf{VV-OPD}
& \verfullcell{15.73}{8.09}
& \verfullcell{42.42}{5.71}
& \verfullcell{16.49}{3.54}
& \verfullcell{36.77}{3.87} \\

\bottomrule
\end{tabular*}
\end{table}

\noindent\textbf{Step-level visual reasoning errors.}
Tab.~\ref{tab:visual_consistency} shows that VV-OPD achieves the lowest visual reasoning error rates across both benchmark groups. Compared with OPD, VV-OPD reduces VER-chain by 3.34 and 2.34 points on Math Reasoning and General Multimodal benchmarks, respectively. The reductions in VER-perc are 3.24 and 5.24 points. These results indicate that the improvements in final-answer accuracy are accompanied by fewer visually contradicted reasoning steps. In particular, the lower perception-step error rates suggest that VV-OPD improves the reliability of visual claims themselves, rather than merely changing the proportion of visual and non-visual steps in the reasoning chain.

\noindent\textbf{Accuracy versus visual faithfulness.}
Final-answer accuracy does not necessarily reflect the correctness of intermediate visual reasoning. For example, VA-OPD improves Math Reasoning Avg@4 over OPD from 33.62 to 36.88, but increases VER-chain from 19.07 to 22.88 and VER-perc from 45.66 to 48.19. A similar pattern appears on General Multimodal benchmarks, where VA-OPD improves Avg@4 from 44.60 to 46.50 while increasing VER-chain from 18.83 to 22.73 and VER-perc from 42.01 to 45.42. These results suggest that strengthening visual supervision does not necessarily improve the correctness of the underlying visual claims. In contrast, VV-OPD achieves higher accuracy while reducing both visual reasoning error rates, highlighting the importance of verifying teacher visual claims rather than simply increasing their influence.

\subsection{Ablation Study}
\label{sec:ablation}


\begin{table}[t]
\centering
\caption{\textbf{Ablation study.}
Effects of visual consistency ranking, edit-token masking,
and reward gating. We report Avg@4 and step-level
visual error rates (VER, \%).
Higher Avg@4 and lower VER are better.}
\label{tab:reward_gating}

\footnotesize
\setlength{\tabcolsep}{1.5pt}
\renewcommand{\arraystretch}{1.04}

\begin{tabular*}{0.80\linewidth}{
@{\extracolsep{\fill}}lcccccc@{}}
\toprule

\multirow{2}{*}{\textbf{Method}}
& \multicolumn{3}{c}{
  \cellcolor{mathbg}\textbf{Math Reasoning}}
& \multicolumn{3}{c}{
  \cellcolor{visbg}\textbf{General Multimodal}} \\
\cmidrule(lr){2-4}
\cmidrule(l){5-7}

& Avg@4 $\uparrow$
& chain $\downarrow$
& perc.\ $\downarrow$
& Avg@4 $\uparrow$
& chain $\downarrow$
& perc.\ $\downarrow$ \\
\midrule

OPD
& 33.62 & 19.07 & 45.66
& 44.60 & 18.83 & 42.01 \\

VV-OPD (w/o mask)
& 35.12 & 18.34 & 44.95
& 45.74 & 18.22 & 40.65 \\

VV-OPD (w/o gating)
& 36.21 & 17.81 & 44.43
& 46.72 & 17.70 & 39.47 \\

\midrule

\textbf{VV-OPD (full)}
& \textbf{40.48} & \textbf{15.73} & \textbf{42.42}
& \textbf{48.97} & \textbf{16.49} & \textbf{36.77} \\

\bottomrule
\end{tabular*}
\end{table}

\noindent Tab.~\ref{tab:reward_gating} examines the contributions of VCR, edit-token masking, and reward gating. Without gating, VCR improves Avg@4 over OPD by 2.59 and 2.12 points on Math Reasoning and General Multimodal benchmarks, respectively. It also reduces VER-chain by 1.26/1.13 points and VER-perc by 1.23/2.54 points, indicating that verified contrastive pairs provide useful corrective supervision even without selective activation. Removing edit-token masking still yields lower visual reasoning error rates than OPD, but produces higher VER than full VV-OPD across all four metrics. This observation is consistent with the benefit of concentrating ranking supervision on the disputed visual content rather than the entire reasoning step. The full model achieves the lowest VER and the highest reported Avg@4.


\subsection{Teacher Scaling}
\label{sec:teacher_scaling}

Tab.~\ref{tab:scale} examines the effect of teacher scale while keeping the 2B student and Geo3K training data fixed. With a 4B teacher, VV-OPD improves over OPD by 6.58 and 3.73 points on Math Reasoning and General Multimodal benchmarks, respectively. The corresponding gains are 6.86 and 4.37 points with an 8B teacher. Notably, VV-OPD with a 4B teacher already outperforms standard OPD with an 8B teacher by 5.17 and 3.24 points.
Scaling the teacher to 32B further improves the absolute performance of both methods. VV-OPD reaches 42.04 and 51.99 Avg@4, exceeding the matched OPD by 3.99 and 3.43 points, respectively. Although the relative gains are smaller than those obtained with the 8B teacher, VV-OPD consistently outperforms OPD across all three teacher scales. At 32B, the improvements extend to all eight individual benchmarks, with particularly large gains on WeMath (+7.50) and MMMU (+5.34). However, the smaller gain on MV-MATH (+0.40) suggests that multi-image reasoning remains challenging even with a stronger teacher. These results indicate that verified visual corrections remain beneficial as teacher capacity increases, while larger teachers also provide stronger distillation targets.

\begin{table}[t]
\centering
\caption{\textbf{OPD vs.\ VV-OPD across teacher scales.}
All models are trained on Geo3K with a 2B student.
Results are reported as Avg@4 (\%), with group averages
computed across four benchmarks.
Green labels indicate improvements over the matched OPD baseline.}
\label{tab:scale}

\footnotesize
\setlength{\tabcolsep}{2pt}
\renewcommand{\arraystretch}{1.10}

\resizebox{\linewidth}{!}{%
\begin{tabular}{@{}ll*{10}{c}@{}}
\toprule

\multirow{2}{*}{\textbf{T$\rightarrow$S}}
& \multirow{2}{*}{\textbf{Method}}
& \multicolumn{5}{c}{
  \cellcolor{mathbg}\textbf{Math Reasoning}}
& \multicolumn{5}{c}{
  \cellcolor{visbg}\textbf{General Multimodal}} \\
\cmidrule(lr){3-7}
\cmidrule(l){8-12}

& & MathVerse & WeMath & MV-MATH & MathVision
& \textbf{Avg}
& LogicVista & MMMU & MMMU-Pro & MMStar
& \textbf{Avg} \\
\midrule

\multirow{2}{*}{4B$\rightarrow$2B}
& OPD
& 33.86 & 56.54 & 18.21 & 20.21 & 32.21
& 40.16 & 43.06 & 34.45 & 58.77 & 44.11 \\

& \textbf{VV-OPD}
& 41.11 & 67.40 & 19.16 & 27.50
& \textbf{38.79}\,\scoreup{6.58}
& 46.42 & 46.97 & 37.08 & 60.87
& \textbf{47.84}\,\scoreup{3.73} \\

\midrule

\multirow{2}{*}{8B$\rightarrow$2B}
& OPD
& 34.59 & 58.94 & 19.79 & 21.16 & 33.62
& 43.96 & 41.67 & 34.03 & 58.72 & 44.60 \\

& \textbf{VV-OPD}
& 42.96 & 69.97 & 20.31 & 28.67
& \textbf{40.48}\,\scoreup{6.86}
& 47.09 & 48.50 & 38.14 & 62.17
& \textbf{48.97}\,\scoreup{4.37} \\

\midrule

\multirow{2}{*}{32B$\rightarrow$2B}
& OPD
& 41.38 & 63.35 & 21.35 & 26.10 & 38.05
& 48.24 & 47.33 & 38.75 & 59.90 & 48.56 \\

& \textbf{VV-OPD}
& 44.80 & 70.85 & 21.75 & 30.75
& \textbf{42.04}\,\scoreup{3.99}
& 50.49 & 52.67 & 41.96 & 62.82
& \textbf{51.99}\,\scoreup{3.43} \\

\bottomrule
\end{tabular}%
}
\end{table}

\subsection{Analysis on Held-out Teacher Errors}
\label{sec:training_dynamics}

We analyze whether VV-OPD reduces the transfer of teacher errors beyond its
training pairs. Both VV-OPD and OPD are trained on Geo3K and evaluated on
held-out DMC data, whose errors are dominated by numeric values rather than
the spatial relations and geometric elements of the training pairs
(Fig.~\ref{fig:correction_types}). We examine the student at two levels: how
it scores fixed erroneous and corrected steps, and whether it repeats teacher
errors when generating freely.

\begin{figure}[t]
  \centering
  \includegraphics[width=.618\linewidth]{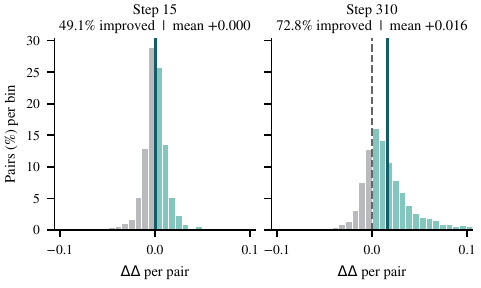}
  \caption{\textbf{Distribution of ranking-margin gains} $\Delta\Delta$ on
  held-out DMC pairs at training steps 15 and 310. The x-axis shows the
  central 98\% of values; statistics use all pairs.}
  \label{fig:training_dynamics}
\end{figure}

\begin{figure}[h]
  \centering
  \includegraphics[width=.618\linewidth]{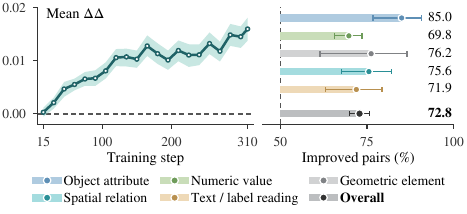}
  
\caption{\textbf{Ranking-margin gains on held-out DMC pairs.}
Left: mean $\Delta\Delta$ across training steps with 95\% confidence intervals.
Right: improved pairs ($\Delta\Delta>0$) at step 310 by correction type,
with Wilson 95\% confidence intervals.}

  \label{fig:margin_types}
\end{figure}

\noindent\textbf{Ranking margins during training.}
For each of 1,000 held-out DMC pairs, we compute the ranking margin
$\Delta = s(z^+) - s(z^-)$, where $s$ averages the student's log-probabilities
over all tokens of the step rather than only the edited tokens as in
Eq.~\eqref{eq:VV-OPD-score}, and the paired gain
$\Delta\Delta = \Delta_{\mathrm{VV\text{-}OPD}} - \Delta_{\mathrm{OPD}}$
between checkpoints at the same training step, so that $\Delta\Delta > 0$
means a larger margin under VV-OPD. The mean gain grows steadily from
approximately zero at step 15 to $+0.016$ at step 310, and its 95\% confidence
interval stays above zero from step 30 onward
(Fig.~\ref{fig:margin_types}, left). The gain is not driven by a few pairs:
the whole distribution shifts toward positive values
(Fig.~\ref{fig:training_dynamics}), and the share of pairs with a positive
gain rises from 49.1\% to 72.8\%. Nor is it driven by a single error type: at
step 310, the share of improved pairs exceeds the no-difference level of 50\%
(dashed line) for every correction type, with Wilson 95\% intervals above
50\% (Fig.~\ref{fig:margin_types}, right). This includes numeric values, which
account for 57\% of the DMC pairs but under 5\% of the Geo3K training pairs.
VV-OPD thus increasingly assigns larger
correction--error margins than OPD, even for error types that are rare in its training pairs.

\begin{figure}[t]
  \centering
  \includegraphics[width=.618\linewidth]{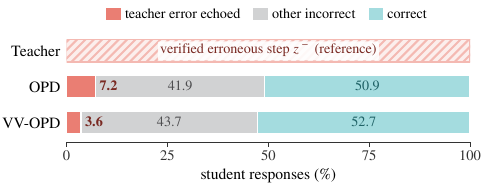}
  \caption{\textbf{Teacher error inheritance} on held-out DMC
  questions with a verified teacher error $z^-$. Each student response either
  echoes the teacher's erroneous claim (lexical match with $z^-$), is otherwise
  incorrect, or is correct.}
  \label{fig:inheritance}
\end{figure}

\noindent\textbf{Teacher error inheritance.}
The margins above are measured on fixed candidate steps. To test whether the
student also repeats the teacher's errors when it generates freely, we take
the held-out DMC questions whose teacher step was verified as erroneous and
classify each student response as echoing the teacher's erroneous claim,
otherwise incorrect, or correct (Fig.~\ref{fig:inheritance}). 
OPD lexically echoes the teacher's erroneous claim in 7.2\% of its responses, compared with 3.6\% for VV-OPD. Meanwhile, the proportion of correct responses increases from 50.9\% to 52.7\%, whereas other incorrect responses increase from 41.9\% to 43.7\%. This suggests that avoiding a particular teacher visual reasoning error does not necessarily resolve all subsequent reasoning failures. Nevertheless, the reduced recurrence of the teacher's erroneous claim provides complementary evidence to the ranking-margin analysis. Because echoes are detected through lexical matching, these rates should be interpreted as surface-form proxies rather than direct measurements of semantic error inheritance.

\subsection{Qualitative Analysis}
\label{sec:qualitative}

\begin{figure}[t]
  \centering
  \includegraphics[width=.618\linewidth]{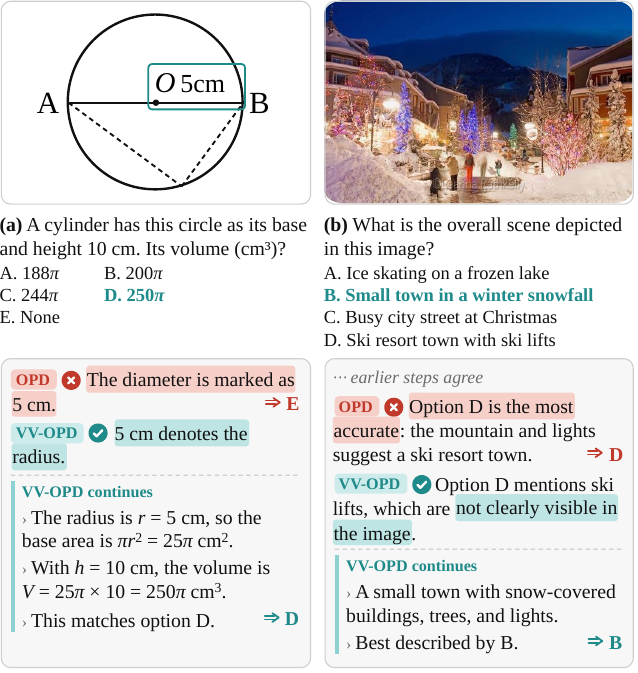}

\caption{\textbf{Qualitative comparisons of OPD and VV-OPD.}
(a) Mathematical reasoning on WeMath.
(b) General multimodal understanding on MMStar.
Red and teal highlight erroneous visual claims and their
image-grounded corrections, respectively.
Full reasoning traces are provided in the Supplementary Material.}

  \label{fig:qualitative}
\end{figure}

Fig.~\ref{fig:qualitative} compares OPD and VV-OPD on two visual reasoning tasks.
\textbf{(a) Mathematical reasoning.}
In the WeMath example, OPD misreads the 5\,cm label on radius $OB$
as the diameter, leading to $r=2.5$ and $V=62.5\pi$.
Although the subsequent calculation is consistent with this
incorrect premise, the final answer is wrong.
VV-OPD correctly identifies the radius and obtains $250\pi$,
yielding the correct answer. The two responses use the same
volume formula but differ in their interpretation of the diagram.
\textbf{(b) General multimodal understanding.}
In Fig.~\ref{fig:qualitative}b, OPD interprets a snow-covered town
as a ski resort, relying on a ski-lift claim not supported by the image.
VV-OPD instead notes that ski lifts are not clearly visible and
describes the buildings, trees, and lights, selecting the correct option.
Across both examples, VV-OPD grounds its visual claims in the image,
whereas OPD makes interpretations unsupported by visual evidence.
Additional qualitative comparisons and complete reasoning traces
for both tasks are provided in the Supplementary Material.

\section{Conclusion}
\label{sec:conclusion}
We identify a failure mode of OPD for VLM reasoning: token-level teacher matching can transfer claims that contradict the image to the student. To address this, we propose VV-OPD, which constructs independently verified contrastive pairs from erroneous teacher reasoning steps and applies a ranking loss to the edited tokens to favor image-grounded corrections. Reward gating focuses this supervision on questions for which the student produces incorrect rollouts. Across eight benchmarks, VV-OPD outperforms OPD and its vision variants on mathematical and general multimodal reasoning tasks while reducing step-level visual reasoning errors. These results show that verifying and correcting teacher claims can improve both final-answer accuracy and the visual grounding of intermediate reasoning. Future work will extend verified corrections to reasoning over multimedia inputs.

\clearpage
\bibliography{main}
\bibliographystyle{plainnat}

\clearpage
\appendix
\setcounter{figure}{0}\setcounter{table}{0}\setcounter{equation}{0}
\renewcommand{\thefigure}{S\arabic{figure}}
\renewcommand{\thetable}{S\arabic{table}}
\renewcommand{\theequation}{S\arabic{equation}}
\renewcommand{\theHfigure}{S\arabic{figure}}
\renewcommand{\theHtable}{S\arabic{table}}
\renewcommand{\theHequation}{S\arabic{equation}}

\pdfbookmark[0]{Supplementary Material}{supplementary}
\begin{center}
  {\LARGE\bfseries Supplementary Material}
\end{center}
\vspace{0.6em}

\section{Implementation Details}
\label{sec:supp_implementation}
 
\subsection{Training Framework and Hyperparameters}
\label{sec:supp_training}

Tab.~\ref{tab:supp_hyperparameters} summarizes the training hyperparameters of VV-OPD, while the following paragraphs describe the implementation details. All training runs use two NVIDIA H200 GPUs (141\,GB memory each)
on a single node.

\begin{table}[!b]
\centering
\caption{\textbf{Training hyperparameters of VV-OPD}
in the primary 8B$\rightarrow$2B setting,
grouped into optimization, on-policy rollout,
and distillation with visual consistency ranking (VCR).}
\label{tab:supp_hyperparameters}

\small
\setlength{\tabcolsep}{5pt}
\renewcommand{\arraystretch}{1.12}

\begin{tabular*}{0.65\linewidth}{
@{\extracolsep{\fill}}ll@{}}
\toprule
\textbf{Hyperparameter} & \textbf{Value} \\
\midrule

\multicolumn{2}{@{}l}{
  \cellcolor{black!5}\textbf{Optimization}} \\
Optimizer & AdamW \\
AdamW $(\beta_1, \beta_2)$ & $(0.9,\,0.999)$ \\
Learning rate & $1\times10^{-6}$ \\
LR schedule & Constant \\
Weight decay & 0.01 \\
Gradient clipping (max.\ norm) & 1.0 \\
Precision & bfloat16 \\
Training epochs & 5 \\
Batch size (questions) & 16 \\
Mini-batch size (questions) & 16 \\

\midrule
\multicolumn{2}{@{}l}{
  \cellcolor{mathbg}\textbf{On-Policy Rollout}} \\
Rollouts per question $|\mathcal{Y}_x|$ & 4 \\
Sampling temperature & 1.0 \\
Top-$p$ & 1.0 \\
Max.\ prompt length & 4,096 \\
Max.\ response length & 4,096 \\

\midrule
\multicolumn{2}{@{}l}{
  \cellcolor{visbg}\textbf{Distillation and VCR}} \\
Distillation loss & Top-$K$ forward KL \\
KL top-$K$ & 64 \\
VCR weight $\lambda$ & 0.001 \\
Ranking sharpness $\beta$ & 2 \\
VCR gate & $\geq$1 incorrect rollout \\
Pairs per active question & 1 \\
Ranking score & Edited tokens only \\

\bottomrule
\end{tabular*}
\end{table}

\noindent\textbf{Models and optimization.}
We implement VV-OPD in the veRL
framework. In our primary setting, Qwen3-VL-8B-Instruct serves as the frozen
teacher and Qwen3-VL-2B-Instruct as the student.
All student
parameters are updated, without LoRA adapters. Training uses bfloat16
precision with Fully Sharded Data Parallel (FSDP) and gradient checkpointing.

\noindent\textbf{On-policy distillation.}
The student is trained on Geo3K questions. For each question, it samples four
responses at temperature 1.0 with top-$p=1.0$, and prompts and responses are
each limited to 4,096 tokens. The frozen teacher evaluates every
student-generated prefix, and the student minimizes the top-$K$ forward KL to
the teacher's next-token distribution with $K=64$ (Eq.~(1) in the main
paper). No policy-gradient loss is used: final-answer correctness only gates
VCR and does not serve as a reward for the student.

\noindent\textbf{Batching and schedule.}
Each training step processes 16 questions, i.e., 64 student responses. The
mini-batch size equals the batch size, so each step performs a single
optimizer update on the rollouts just sampled. We train for five epochs with
a constant learning rate and no warmup.

\noindent\textbf{Visual consistency ranking.}
VCR uses one verified contrastive pair per active question; the procedure is
detailed in Sec.~\ref{sec:supp_algorithm}.

\subsection{Core Training Algorithm}
\label{sec:supp_algorithm}

Algorithm~\ref{alg:supp_vvopd} summarizes one VV-OPD training step;
equation numbers refer to the main paper. Verified contrastive pairs are
constructed and verified offline (Sec.~3.2 in the main paper). During
training, each question $x$ in the batch is matched to its verified
contrastive pairs $\mathcal{P}_x$, and the reward gate $g(x,\mathcal{Y}_x)$
activates VCR only when at least one of its rollouts yields an incorrect
final answer. The VCR loss is added to the OPD loss with weight $\lambda$,
and both update the same student parameters.

\begin{figure}[!b]
\centering
\begin{minipage}{0.8\linewidth}
\begin{algorithm}[H]
\caption{VV-OPD training with reward-gated VCR}
\label{alg:supp_vvopd}
\small
\begin{algorithmic}[1]
\Require student $\pi_\theta$, frozen teacher $\pi_T$, training set
  $\mathcal{D}$, verified contrastive pairs $\mathcal{P}$, VCR weight
  $\lambda$, ranking sharpness $\beta$
\For{each training step}
  \State Sample a batch of questions $\mathcal{B}\subset\mathcal{D}$
  \For{each $x=(I,q)\in\mathcal{B}$}
    \State Sample rollouts $\mathcal{Y}_x\sim\pi_\theta(\cdot\mid x)$
    \State Evaluate final-answer correctness $r(x,y)$, $y\in\mathcal{Y}_x$
    \State $\ell_x\gets 0$
    \If{$g(x,\mathcal{Y}_x)=1$} \Comment{reward gate, Eq.~(7)}
      \State Sample a pair $p_x=(I,q,h,z^+,z^-)\in\mathcal{P}_x$
      \State $\Delta_x\gets s_\theta(z^+)-s_\theta(z^-)$
        \Comment{edited tokens, Eq.~(5)}
      \State $\ell_x\gets-\log\sigma(\beta\,\Delta_x)$ \Comment{Eq.~(6)}
    \EndIf
  \EndFor
  \State $\mathcal{L}_{\mathrm{OPD}}\gets$ OPD loss on all rollouts
    \Comment{Eq.~(1)}
  \State $\mathcal{L}_{\mathrm{VCR}}\gets\frac{1}{|\mathcal{B}|}
    \sum_{x\in\mathcal{B}}\ell_x$
  \State Take an AdamW step on
    $\mathcal{L}_{\mathrm{OPD}}+\lambda\,\mathcal{L}_{\mathrm{VCR}}$
    \Comment{Eq.~(8)}
\EndFor
\State \Return $\pi_\theta$
\end{algorithmic}
\end{algorithm}
\end{minipage}
\end{figure}

\noindent\textbf{Implementation details.}
$\mathcal{L}_{\mathrm{OPD}}$ is the top-$K$ forward KL from the teacher to
the student, evaluated at every prefix of the student's rollouts. The score
$s_\theta(z^\pm)$ averages the student's log-probabilities over the edited
token positions only, while the full sequence $(I,q,h,z^\pm)$ is used for
conditioning. When a question has several verified contrastive pairs, one of
them is sampled at each update. VCR is reference-free: it requires no
teacher likelihoods for the candidate steps, and neither the teacher nor an
external verifier scores the pairs during training.

\subsection{Prompt Template for Training and Inference}
\label{sec:supp_prompts}

Training uses the step-structured prompt below, given as a single user
message without a system prompt. The same input is given to the student when
it generates rollouts and to the teacher when it scores the student's
prefixes. Evaluation uses the same instruction; for questions with answer
options, the input additionally asks for a single option letter as the final
boxed answer (e.g., ``The final boxed answer must be exactly one option
letter: A, B, C, or D.''). The step tags also delimit the steps used in the
step-level evaluation (Sec.~\ref{sec:supp_eval}).

\begin{tcolorbox}[
  enhanced,
  width=\columnwidth,
  colback=white,
  colframe=black!20,
  colbacktitle=mathbg,
  coltitle=black,
  fonttitle=\bfseries,
  title={Step-Structured Prompt},
  fontupper=\small,
  boxrule=0.6pt,
  arc=1.5mm,
  left=8pt,
  right=8pt,
  top=5pt,
  bottom=5pt,
  before skip=7pt,
  after skip=7pt
]
\raggedright
\textcolor{promptteal}{\texttt{\textless image\textgreater}}\\
\textcolor{promptteal}{\texttt{\{question\}}}\par\smallskip
Solve the problem step by step. Enclose each reasoning step in
\texttt{\textless step\textgreater} and \texttt{\textless/step\textgreater}
tags. Put your final answer in \texttt{\textbackslash boxed\{\}}.
\end{tcolorbox}

\noindent Here, \textcolor{promptteal}{\texttt{\textless image\textgreater}}
and \textcolor{promptteal}{\texttt{\{question\}}} are placeholders for the
input image and the question text, including the answer options; for
questions with interleaved images, the image placeholders appear at their
positions in the question.

\subsection{Baseline Implementation Details}
\label{sec:supp_baselines}

All baselines in the main paper use Qwen3-VL-2B-Instruct as the student
and, where a teacher is required, Qwen3-VL-8B-Instruct as the teacher.
Unless noted otherwise, they are trained on Geo3K.
For SFT~\cite{ouyang2022training}, GRPO~\cite{shao2024deepseekmath},
OPD~\cite{agarwal2024policy}, and VA-OPD~\cite{liu2026visual}, we use the
publicly released implementations with their default configurations.
Since no official implementation of VOLD~\cite{bousselham2026vold} is
available, we reimplement it following the configuration specified in the
original paper, which combines GRPO with on-policy distillation and applies
teacher supervision only to incorrect responses. We replace its original
text-only teacher with Qwen3-VL-8B-Instruct so that it shares the teacher
used by the other methods.
For Vision-OPD~\cite{yuan2026vision}, we use the official training pipeline
and its original training data without modification, except that the
teacher and student are replaced by Qwen3-VL-8B-Instruct and
Qwen3-VL-2B-Instruct, respectively.

\subsection{Evaluation Protocols}
\label{sec:supp_eval}

\noindent\textbf{Answer accuracy.}
We follow the benchmark-specific evaluation protocols implemented in
VLMEvalKit, using an LLM judge (GPT-4o) for answer matching where required.
For each question, we sample four responses at temperature 1.0 with a
maximum generation length of 8,192 tokens and report their mean final-answer
accuracy (Avg@4).

\noindent\textbf{Visual reasoning errors.}
For each benchmark and model, we evaluate 1,000 responses: the first two of
the four responses generated for the accuracy evaluation, on a fixed random
subset of 500 questions shared across all models. Accuracy and visual
consistency are thus measured on the same generations without additional
sampling. For LogicVista, which contains only 447 questions, we use all
questions, yielding 894 responses. Each response is split into reasoning
steps by its \texttt{\textless step\textgreater} tags; responses without step
tags are segmented by GPT-5-mini. GPT-6.1-Terra then serves as a multimodal
judge using the prompt provided below. Given the
original image, the question, and the numbered steps of a response, the judge
extracts every direct perception claim as an exact quote from its step and
labels it as supported, contradicted, or uncertain with respect to the image.
A step belongs to $\mathcal{Z}_{\text{perc}}$ in Eq.~(9) of the main paper if
it contains at least one perception claim, and to $\mathcal{Z}_{\text{err}}$
if at least one of its claims is contradicted. Mathematical deductions,
calculation errors, and final answers are not treated as visual claims.
Results are macro-averaged across benchmarks.

\par\medskip

\begin{tcolorbox}[
  enhanced,
  width=\linewidth,
  colback=white,
  colframe=black!20,
  colbacktitle=mathbg,
  coltitle=black,
  fonttitle=\bfseries,
  title={Visual Reasoning Error Judge Prompt},
  fontupper=\small,
  fontlower=\small,
  segmentation style={black!25, dashed},
  boxrule=0.6pt,
  arc=1.5mm,
  left=8pt,
  right=8pt,
  top=5pt,
  bottom=5pt,
  middle=4pt,
  before skip=8pt,
  after skip=8pt
]

\phantomsection
\label{box:supp_judge_prompt}

\setlength{\parskip}{2.5pt}
\setlength{\parindent}{0pt}
You are an independent annotator of visual hallucinations in multimodal
mathematical reasoning. Inspect the ORIGINAL IMAGE and QUESTION; then identify
the reasoning steps that contain DIRECT PERCEPTION claims grounded in the image.

A claim about a printed number, marked angle, label, tick mark, shape, color,
location, chart value, count, or explicitly depicted spatial relation is
visual. A mathematical deduction, a guessed value that is explicitly marked as
a hypothesis, an algebraic manipulation, an incorrect calculation, or a final
answer is NOT by itself a visual hallucination. Do not use the reference
answer or whether the final answer is correct.

Read and inspect the supplied ORIGINAL IMAGE independently. CRITICAL:
distinguish angle labels (e.g.\ 55\textdegree) from segment-length labels
(e.g.\ x+4 without the degree sign). Do not invent angle measures or visible
numbers absent from the image. The fact that a statement is plausible or
useful for the final solution does NOT make it visually supported.

For each substantive image-perception claim, quote an EXACT contiguous
substring of the given STEP. Never paraphrase, normalize, shorten with
ellipses, change math notation, or quote from outside the step. Step indices
are 1-based: the first is STEP 1, not 0. If you cannot locate an exact quote,
do not invent one. Use `uncertain' if the image itself is ambiguous. For
contradicted claims, give evidence grounded only in the supplied image; when a
mark or label is absent, explicitly say it is not visible rather than
fabricating corrected values.

One reasoning step can contain several claims. A step with any contradicted
claim is ONE hallucinated step. Mark uncertain independently; do not infer
correctness from an absence of claims. Include all substantial direct
perceptual claims; for responses with no directly visual claims, return an
EMPTY claims array. Return only JSON matching the schema.
\tcblower
\raggedright
\textbf{Output schema (JSON).} All fields are required; no other fields are
allowed. \texttt{claims}: list of objects with \texttt{step\_index} (integer,
1-based), \texttt{quote} (exact substring of the step), \texttt{verdict}
(\texttt{supported}\,$|$\,\texttt{contradicted}\,$|$\,\texttt{uncertain}),
\texttt{image\_evidence} (string), and \texttt{corrected\_fact} (string);
\texttt{notes}: string.
\end{tcolorbox}

\section{Computational Cost Overhead}
\label{sec:supp_cost}

\noindent\textbf{Training-time overhead.}
We compare the per-step training time of VV-OPD and standard OPD in the
8B$\rightarrow$2B setting on Geo3K, using identical training configurations
and hardware (Sec.~\ref{sec:supp_training}).
Both runs contain 840 aligned global steps. Without controlling for token
counts, VV-OPD increases the average step time from 39.78\,s to 44.01\,s,
an additional 4.23\,s per step.
In Tab.~\ref{tab:supp_cost_token_matching}, matched steps (\%) denote
the proportion of aligned steps retained at each matching tolerance.

\noindent\textbf{Token-matched comparison.}
Since the number of generated tokens varies across on-policy rollouts,
differences in step time may partly reflect differences in token counts. We
therefore compare the two methods at the same global steps while restricting
their token-count differences. For step $i$, let $N_{\mathrm{OPD}}^{(i)}$ and
$N_{\mathrm{VV}}^{(i)}$ denote the generated token counts. We define the
relative difference and the matched-step set as
\begin{equation}
\label{eq:supp_token_match}
d_i =
\frac{|N_{\mathrm{VV}}^{(i)}-N_{\mathrm{OPD}}^{(i)}|}
{N_{\mathrm{OPD}}^{(i)}},
\qquad
\mathcal{S}_{\tau}=\{i:d_i\leq\tau\},
\end{equation}
where $\tau$ is the matching tolerance. For the retained steps, we compare the
mean step times $\overline{T}_{\mathrm{OPD},\tau}$ and
$\overline{T}_{\mathrm{VV},\tau}$ over the same matched-step set
$\mathcal{S}_{\tau}$.

\noindent As shown in Tab.~\ref{tab:supp_cost_token_matching}, the $\pm$10\%
criterion retains 67.1\% of the aligned training steps. Under this
condition, \mbox{VV-OPD} takes 43.76\,s per step compared with 40.86\,s
for OPD, an additional 2.90\,s (Fig.~\ref{fig:supp_cost_comparison}), or
$1.07\times$ the per-step time of OPD. Across matching tolerances of
5\%--20\%, the additional time ranges from 2.27\,s to 4.03\,s per step,
and it is 4.23\,s without matching; the difference decreases as the
matching becomes stricter. These differences are small relative to a
per-step time of about 40\,s. The matched estimates remain conditional on
the selected steps and do not isolate every source of computational
variation.

\noindent\textbf{Inference cost.}
VV-OPD introduces no additional inference-time modules, parameters, or
forward passes. Verified contrastive pairs and reward gating are used only
during training. At deployment, the student operates without the teacher or
visual verifier. Therefore, VV-OPD incurs no additional model-level inference
computation for a fixed input and output token budget.

\medskip
\noindent
\begin{minipage}[c]{0.51\linewidth}
\centering

\captionof{table}{\textbf{Training-time overhead under
token matching.}
The $\pm$10\% setting is highlighted.}
\label{tab:supp_cost_token_matching}

\vspace{5pt}
\footnotesize
\setlength{\tabcolsep}{3.5pt}
\renewcommand{\arraystretch}{1.08}

\begin{tabular}{@{}cccc@{}}
\toprule
& & \multicolumn{2}{c}{Time (s/step)} \\
\cmidrule(l){3-4}
Token matching & Matched (\%) & OPD & VV-OPD \\
\midrule
$\pm$5\%  & 36.2  & 43.19 & 45.46 \\
\rowcolor{vvteal!35}
$\pm$10\% & 67.1  & 40.86 & 43.76 \\
$\pm$15\% & 87.3  & 40.33 & 43.95 \\
$\pm$20\% & 95.8  & 40.10 & 44.13 \\
\midrule
None (all steps) & 100.0 & 39.78 & 44.01 \\
\bottomrule
\end{tabular}
\end{minipage}%
\hfill
\raisebox{-10pt}{%
\begin{minipage}[c]{0.47\linewidth}
\centering

\includegraphics[width=\linewidth]{
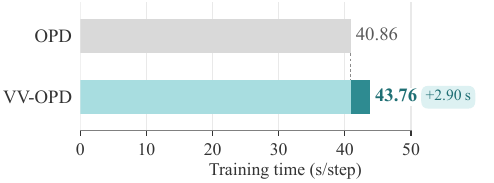}

\vspace{4pt}
\captionof{figure}{\textbf{Training-time comparison.}
Under $\pm$10\% token matching, VV-OPD
requires an additional 2.90\,s per step.}
\label{fig:supp_cost_comparison}

\end{minipage}%
}

\par\medskip

\section{Additional Experimental Results}
\label{sec:suppl_results}

\subsection{Direct Pair-Injection Baseline}
\label{sec:supp_pair_injection}

To distinguish the benefit of explicit visual preference supervision from
simply exposing the student to additional training information, we introduce
a direct pair-injection baseline. Instead of optimizing the VCR ranking
objective, we incorporate verified contrastive pairs directly into the
training prompts.

\noindent Specifically, we first construct a frozen question-level gate using
four rollouts from the initial student model. For each training question
associated with verified contrastive pairs, the gate is activated if at least
one rollout produces an incorrect final answer. We then augment the original
prompt with the corresponding teacher-generated erroneous steps ($z^-$) and
verified visual corrections ($z^+$) for gated questions, subject to the
prompt-length budget. Questions without an activated gate retain their
original prompts.

\noindent The resulting baseline is trained using standard forward-KL
on-policy distillation, with both the VCR loss and the reward-gating branch
disabled. We use the same training questions, student and teacher models,
rollout count, batch size, learning rate, and training steps as VV-OPD.
No additional training examples are introduced.

\noindent As shown in Tab.~\ref{tab:supp_pair_injection}, injecting the
verified contrastive pairs into the prompts leaves performance close to OPD.
Accuracy changes by $+0.37$ and $-0.85$ points, and the visual reasoning error rates
change by at most 1.31 points without a consistent direction: VER-chain
increases slightly ($+1.31$ and $+0.52$), whereas VER-perc decreases
slightly ($-0.37$ and $-1.17$). In contrast, VV-OPD improves accuracy by
6.86 and 4.37 points and reduces all four error rates by 2.34 to 5.24
points. This suggests that the gains come from training the student to rank
the corrected steps above the erroneous ones, rather than from exposure to
the verified contrastive pairs.

\begin{table}[h]
\centering
\caption{\textbf{Direct pair injection versus VCR.}
All methods use the same training questions and
optimization budget. Avg@4 (\%) is macro-averaged
over four benchmarks per group, while visual
reasoning error rates (VER, \%) are averaged over
two benchmarks per group.
Higher Avg@4 and lower VER are better.}
\label{tab:supp_pair_injection}

\footnotesize
\setlength{\tabcolsep}{3pt}
\renewcommand{\arraystretch}{1.10}

\begin{tabular*}{0.8\linewidth}{
@{\extracolsep{\fill}}lcccccc@{}}
\toprule

\multirow{2}{*}{\textbf{Method}}
& \multicolumn{3}{c}{
  \cellcolor{mathbg}\textbf{Math Reasoning}}
& \multicolumn{3}{c}{
  \cellcolor{visbg}\textbf{General Multimodal}} \\
\cmidrule(lr){2-4}
\cmidrule(l){5-7}

& Avg@4 $\uparrow$
& chain $\downarrow$
& perc.\ $\downarrow$
& Avg@4 $\uparrow$
& chain $\downarrow$
& perc.\ $\downarrow$ \\
\midrule

OPD
& 33.62 & 19.07 & 45.66
& 44.60 & 18.83 & 42.01 \\

OPD + pair injection
& 33.99 & 20.38 & 45.29
& 43.75 & 19.35 & 40.84 \\

\midrule

\textbf{VV-OPD (ours)}
& \textbf{40.48} & \textbf{15.73} & \textbf{42.42}
& \textbf{48.97} & \textbf{16.49} & \textbf{36.77} \\

\bottomrule
\end{tabular*}
\end{table}

\subsection{Sensitivity to VCR Hyperparameters}
\label{sec:supp_vcr_sensitivity}

\begin{figure}[h]
\centering
\includegraphics[width=.618\linewidth]{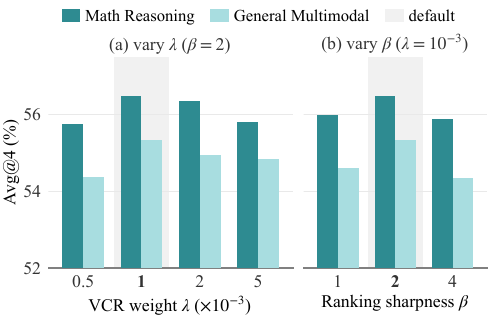}
\caption{\textbf{Sensitivity to VCR hyperparameters.}
Avg@4 when varying (a) the VCR weight $\lambda$ at $\beta=2$ and (b) the
ranking sharpness $\beta$ at $\lambda=10^{-3}$. Math and General denote the
mean over MathVerse and WeMath and over MMMU and MMStar, respectively. The
shaded region marks the default setting.}
\label{fig:supp_vcr_sensitivity}
\end{figure}

\noindent We study the two hyperparameters of VCR: the loss weight $\lambda$ and the
ranking sharpness $\beta$ (Eqs.~(6) and~(8) in the main paper). Starting from
the default setting ($\lambda=10^{-3}$, $\beta=2$), we vary
$\lambda\in\{0.5,1,2,5\}\times10^{-3}$ with $\beta=2$ and
$\beta\in\{1,2,4\}$ with $\lambda=10^{-3}$, and additionally increase both
jointly ($\lambda=2\times10^{-3}$, $\beta=4$). All other settings follow
Tab.~\ref{tab:supp_hyperparameters}. For computational efficiency, all
settings, including OPD ($\lambda=0$) and the default, are evaluated on
MathVerse and WeMath for mathematical reasoning and MMMU and MMStar for
general multimodal reasoning, and we report Avg@4 averaged within each group.
Fig.~\ref{fig:supp_vcr_sensitivity} plots the group averages along each
sweep, and Tab.~\ref{tab:supp_vcr_sensitivity} lists all settings.

\begin{table}[t]
\centering
\caption{\textbf{Sensitivity to VCR hyperparameters.}
Rows vary $\lambda$ (with $\beta=2$), then $\beta$
(with $\lambda=10^{-3}$), then both jointly.
All scores are Avg@4 (\%); Avg.\ is the mean
of the two benchmarks in each group.
The default setting is highlighted.}
\label{tab:supp_vcr_sensitivity}

\footnotesize
\setlength{\tabcolsep}{3pt}
\renewcommand{\arraystretch}{1.10}

\begin{tabularx}{0.8\linewidth}{
@{}cc*{6}{>{\centering\arraybackslash}X}@{}}
\toprule

\multirow{2}{*}{
  \shortstack{$\lambda$\\[-1pt]
  {\scriptsize($\times10^{-3}$)}}}
& \multirow{2}{*}{$\beta$}
& \multicolumn{3}{c}{
  \cellcolor{mathbg}\textbf{Math Reasoning}}
& \multicolumn{3}{c}{
  \cellcolor{visbg}\textbf{General Multimodal}} \\
\cmidrule(lr){3-5}
\cmidrule(l){6-8}

& & MathVerse & WeMath & \textbf{Avg.}
  & MMMU & MMStar & \textbf{Avg.} \\
\midrule

\multicolumn{2}{c}{OPD ($\lambda=0$)}
& 34.59 & 58.94 & 46.77
& 41.67 & 58.72 & 50.20 \\

\midrule

0.5 & 2
& 42.10 & 69.41 & 55.76
& 47.81 & 60.93 & 54.37 \\

\rowcolor{vvteal!35}
1 & 2
& 42.96 & 69.97 & \textbf{56.47}
& 48.50 & 62.17 & \textbf{55.34} \\

2 & 2
& 42.97 & 69.73 & 56.35
& 48.42 & 61.48 & 54.95 \\

5 & 2
& 42.38 & 69.22 & 55.80
& 48.72 & 60.95 & 54.84 \\

\midrule

1 & 1
& 42.39 & 69.58 & 55.99
& 47.67 & 61.55 & 54.61 \\

1 & 4
& 42.35 & 69.41 & 55.88
& 47.50 & 61.20 & 54.35 \\

\midrule

2 & 4
& 41.76 & 69.11 & 55.44
& 47.17 & 61.57 & 54.37 \\

\bottomrule
\end{tabularx}
\end{table}

\noindent VV-OPD is robust to both hyperparameters. Every evaluated setting
improves over OPD on this subset by at least 8.67 points on mathematical
reasoning and 4.15 points on general multimodal reasoning, compared with
gains of 9.70 and 5.14 points at the default setting. Across a tenfold range
of $\lambda$, the two group averages vary by at most 0.71 and 0.97 points,
respectively. Doubling $\lambda$ changes them little ($-0.12$ and $-0.39$),
while both ends of the range lead to small decreases, with the largest at
$\lambda=0.5\times10^{-3}$ ($-0.71$ and $-0.97$). Changing $\beta$ in either
direction also lowers both averages slightly, by 0.48 and 0.73 points for a
softer ranking ($\beta=1$) and by 0.59 and 0.99 points for a sharper one
($\beta=4$). Increasing $\lambda$ and $\beta$ jointly gives the lowest
mathematical-reasoning average among the evaluated settings (55.44),
suggesting that a stronger and sharper ranking signal does not bring further
gains. Overall, the default setting achieves the highest average in both
groups, and all other settings remain within 1.03 and 0.99 points of it. On
individual benchmarks, other settings are occasionally better, but by at most
0.22 points (MMMU at $\lambda=5\times10^{-3}$).

\section{Verified Contrastive Pairs}
\label{sec:suppl_pairs}

\subsection{Construction and Coverage}
\label{sec:supp_pair_coverage}

\noindent\textbf{Construction.}
Verified contrastive pairs are constructed in three stages
(Sec.~3.2 of the main paper). Their key
instructions are summarized below.

\begin{tcolorbox}[
  enhanced,
  width=\columnwidth,
  colback=white,
  colframe=black!20,
  colbacktitle=mathbg,
  coltitle=black,
  fonttitle=\bfseries,
  title={Pair Construction Prompts (Key Instructions)},
  fontupper=\small,
  boxrule=0.6pt,
  arc=1.5mm,
  left=7pt,
  right=7pt,
  top=3pt,
  bottom=3pt,
  before skip=4pt,
  after skip=9pt
]
\setlength{\parindent}{0pt}
\setlength{\parskip}{2pt}
\setlength{\emergencystretch}{1em}

\textbf{(1) Detection.}
Identify the earliest independent visual claim
contradicted by the image, with a clean preceding
prefix. Exclude math-only errors, downstream
inferences, and uncertain visual observations.

\textbf{(2) Local correction.}
Minimally correct the contradicted visual claim
within the original teacher step. Preserve the
preceding context and all unaffected text;
do not add reasoning steps or alter other claims.

\textbf{(3) Independent verification.}
Retain the pair only if the original claim
contradicts the image, the correction is visually
supported, and the preceding prefix is clean.
Require a minimal same-fact edit, preservation
of unaffected text, and exact-quoted spans.
Reject uncertain cases or newly introduced errors.
\end{tcolorbox}

\begin{figure}[h]
  \centering
  \includegraphics[width=.60\linewidth]{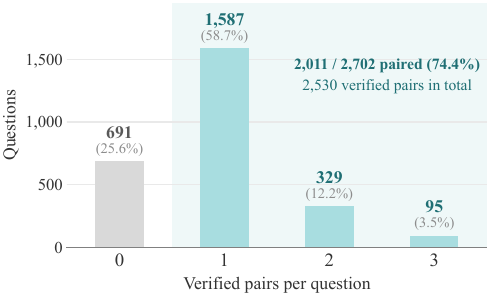}
  \caption{\textbf{Distribution of verified contrastive pairs per Geo3K
  question.} Of 2,702 audited questions, 2,011 (74.4\%) have at least one
  verified contrastive pair, yielding 2,530 pairs in total. Bar labels show
  question counts and percentages of all audited questions.}
  \label{fig:pair_distribution}
\end{figure}

\noindent\textbf{Coverage.}
We report coverage on Geo3K, the training set from which all verified
contrastive pairs used by VCR are drawn. We audit 2,702 Geo3K questions with
step-structured teacher trajectories and obtain 2,530 verified contrastive
pairs across 2,011 questions (74.4\% coverage). As shown in
Fig.~\ref{fig:pair_distribution}, 1,587 questions yield one pair, 329 yield
two, and 95 yield three; the remaining 691 have no verified contrastive pair.
Among questions with at least one pair, 78.9\% contribute exactly one pair.
The pairs from the DVQA--M3CoT Collection (DMC; Sec.~4.1 in the main paper)
are constructed for held-out analysis only (Sec.~4.6 in the main paper) and
are therefore not included in these statistics.

\subsection{Examples of Verified Contrastive Pairs}
\label{sec:supp_pair_examples}
\begin{figure}[htbp]
  \centering
  \includegraphics[width=1.0\linewidth]{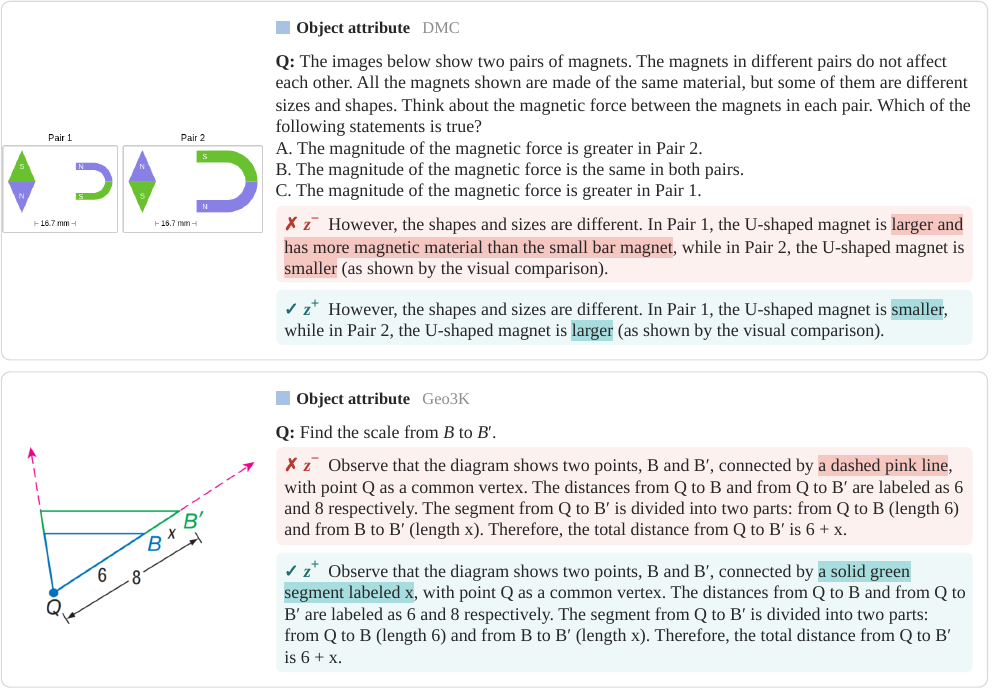}
  \caption{\textbf{Verified contrastive pairs (1/3): object attribute.}
  Each example shows the input image and question, the teacher's erroneous
  step $z^-$, and its verified correction $z^+$ under the same preceding
  context. Highlighted spans are the edited tokens scored by VCR (Eq.~(5) in
  the main paper). The source dataset is marked on each example; DMC pairs
  are held out from training.}
  \label{fig:supp_pairs_1}
\end{figure}

\begin{figure}[tp]
  \centering
  \includegraphics[width=1.0\linewidth]{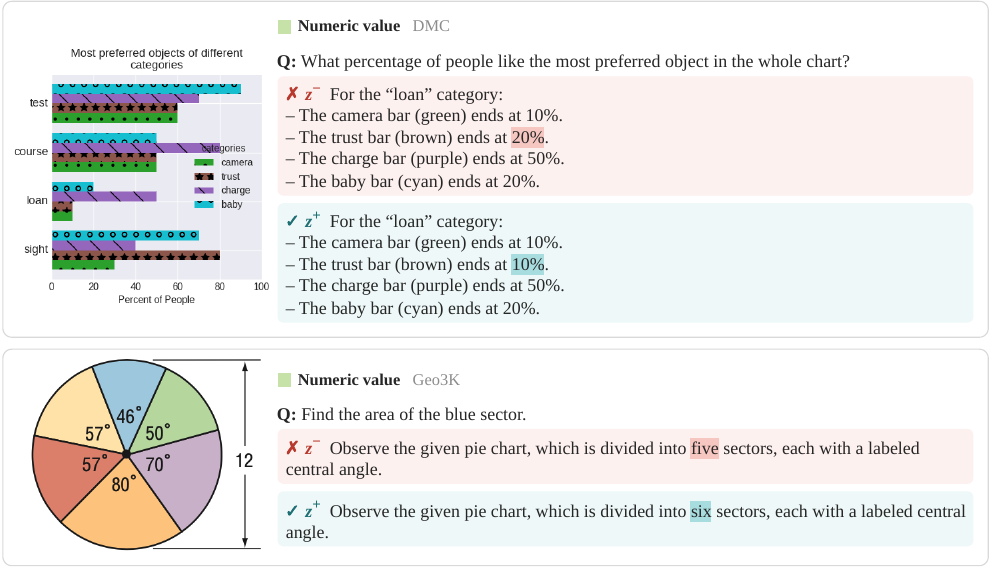}\par\vspace{10pt}
  \includegraphics[width=1.0\linewidth]{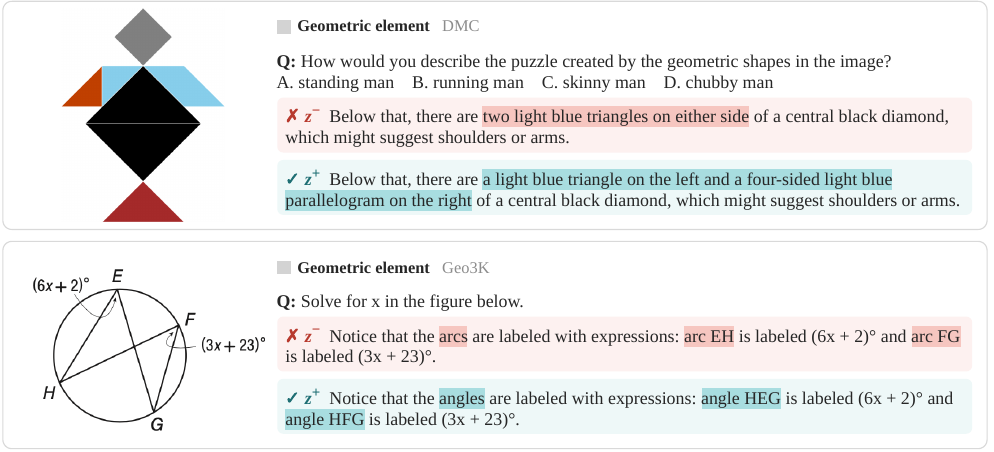}
  \caption{\textbf{Verified contrastive pairs (2/3): numeric value and
  geometric element.} Notation as in Fig.~\ref{fig:supp_pairs_1}.}
  \label{fig:supp_pairs_2}
\end{figure}

\begin{figure}[tp]
  \centering
  \includegraphics[width=1.0\linewidth]{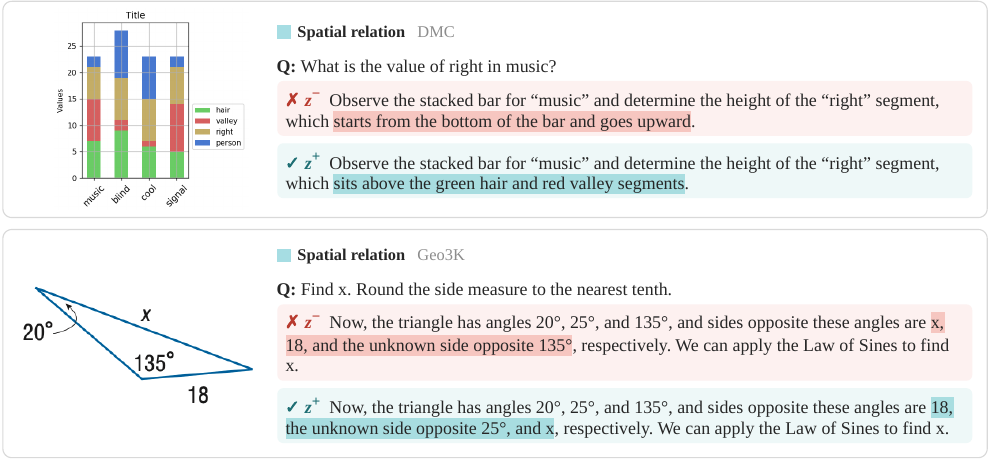}\par\vspace{10pt}
  \includegraphics[width=1.0\linewidth]{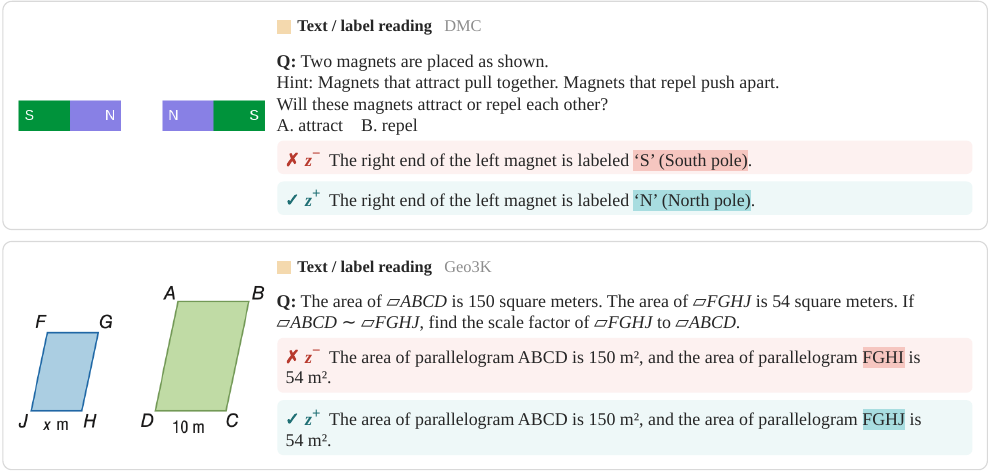}
  \caption{\textbf{Verified contrastive pairs (3/3): spatial relation and
  text\,/\,label reading.} Notation as in Fig.~\ref{fig:supp_pairs_1}.}
  \label{fig:supp_pairs_3}
\end{figure}

Figs.~\ref{fig:supp_pairs_1}--\ref{fig:supp_pairs_3} show two selected pairs
for each of the five correction types in Fig.~3 of the main paper. Unlike the
coverage statistics above, the examples are drawn from both Geo3K and DMC,
and the source of each example is marked next to its correction type. Geo3K
pairs are used by VCR during training, whereas DMC pairs are held out from
training. We include DMC because its errors cover types that are rare in
Geo3K: numeric values account for 57\% of the DMC pairs but under 5\% of the
Geo3K pairs (Sec.~4.6 in the main paper). Each example shows the input image
and question, the teacher's erroneous step $z^-$, and its verified correction
$z^+$ under the same preceding context. Highlighted spans are the edited
tokens scored by VCR; the rest of the step is shared by $z^-$ and $z^+$. 
The examples also make the relevant visual evidence
explicit, allowing readers to inspect why $z^-$
contradicts the image and $z^+$ is visually supported.

\makeatletter
\setlength{\@fptop}{0pt}
\setlength{\@fpsep}{12pt plus 2pt}
\setlength{\@fpbot}{0pt plus 1fil}
\makeatother

\clearpage

\begin{figure}[t]
  \centering
  \includegraphics[width=\linewidth]{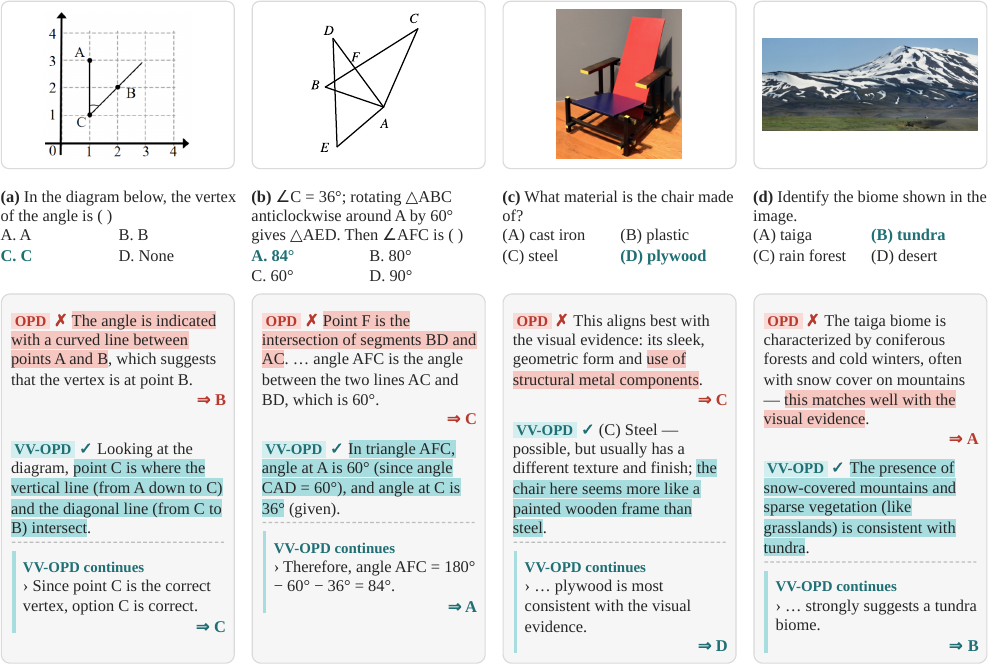}
  \caption{\textbf{Additional qualitative comparisons of OPD and
  VV-OPD.} (a, b) Mathematical reasoning on WeMath and MathVerse. (c, d)
  General multimodal understanding on MMMU. Red and teal highlight erroneous
  visual claims and their image-grounded corrections, respectively. Full
  reasoning traces are provided in
  Figs.~\ref{fig:supp_trace_angle}--\ref{fig:supp_trace_biome}.}
  \label{fig:supp_qual_examples}
\end{figure}

\section{Additional Qualitative Comparisons}
\label{sec:suppl_qualitative}

\noindent\textbf{Additional examples.}
Fig.~\ref{fig:supp_qual_examples} shows four additional examples:
two on mathematical reasoning (WeMath and MathVerse) and two on
general multimodal understanding (MMMU).
In (a), OPD attributes the angle mark to points A and B and selects
B as the vertex, whereas VV-OPD identifies C where the two rays meet.
In (b), OPD takes F to be the intersection of BD and AC and equates
$\angle AFC$ with the $60^\circ$ rotation angle, whereas VV-OPD computes
$\angle AFC$ in triangle AFC from $\angle CAD=60^\circ$ and
$\angle C=36^\circ$.
In (c), OPD asserts that the chair uses structural metal components
and selects steel, whereas VV-OPD describes a painted wooden frame
and selects plywood. In (d), OPD claims that the coniferous forests
of the taiga match the image, which shows no trees, whereas VV-OPD
relies on the snow-covered mountains and sparse, grassy vegetation
and selects tundra.

\begin{figure*}[p]
  \centering
  \includegraphics[width=1.0\textwidth]{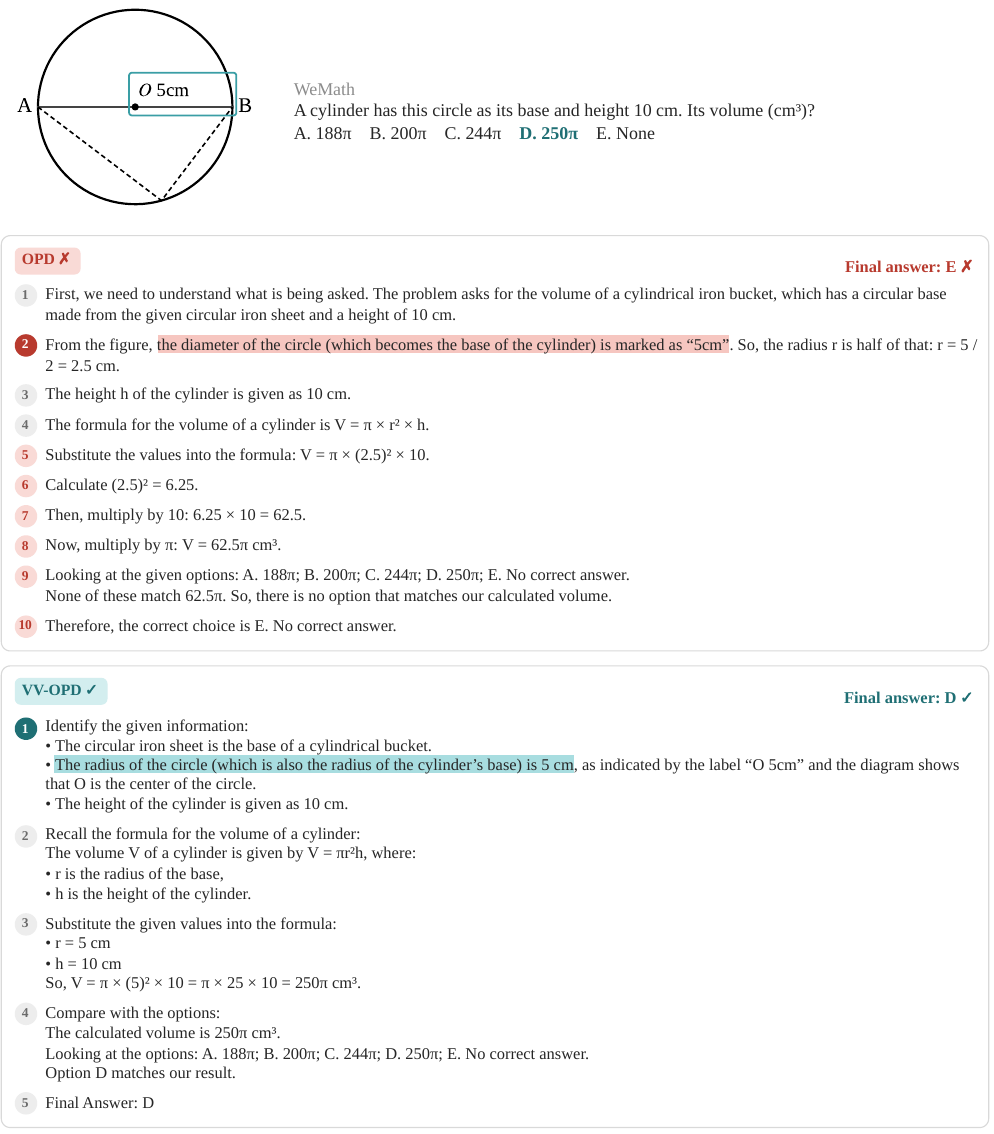}
  \caption{\textbf{Full reasoning traces for Fig.~7(a) in the main paper}
  (WeMath). Dark badges mark the step containing the disputed visual claim,
  and light red badges mark OPD steps that build on its error.}
  \label{fig:supp_trace_cylinder}
\end{figure*}

\begin{figure*}[p]
  \centering
  \includegraphics[width=1.0\textwidth]{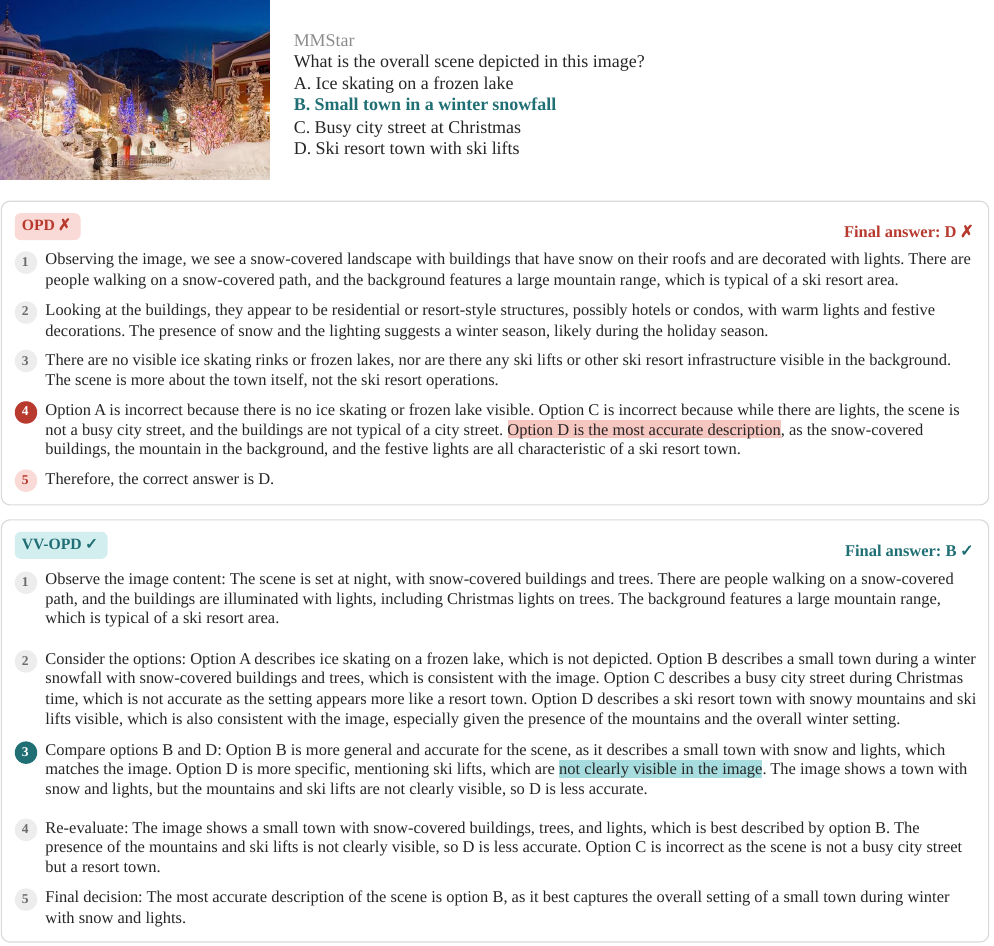}
  \caption{\textbf{Full reasoning traces for Fig.~7(b) in the main paper}
  (MMStar). Notation as in Fig.~\ref{fig:supp_trace_cylinder}.}
  \label{fig:supp_trace_town}
\end{figure*}

\begin{figure*}[p]
  \centering
  \includegraphics[width=1.0\textwidth]{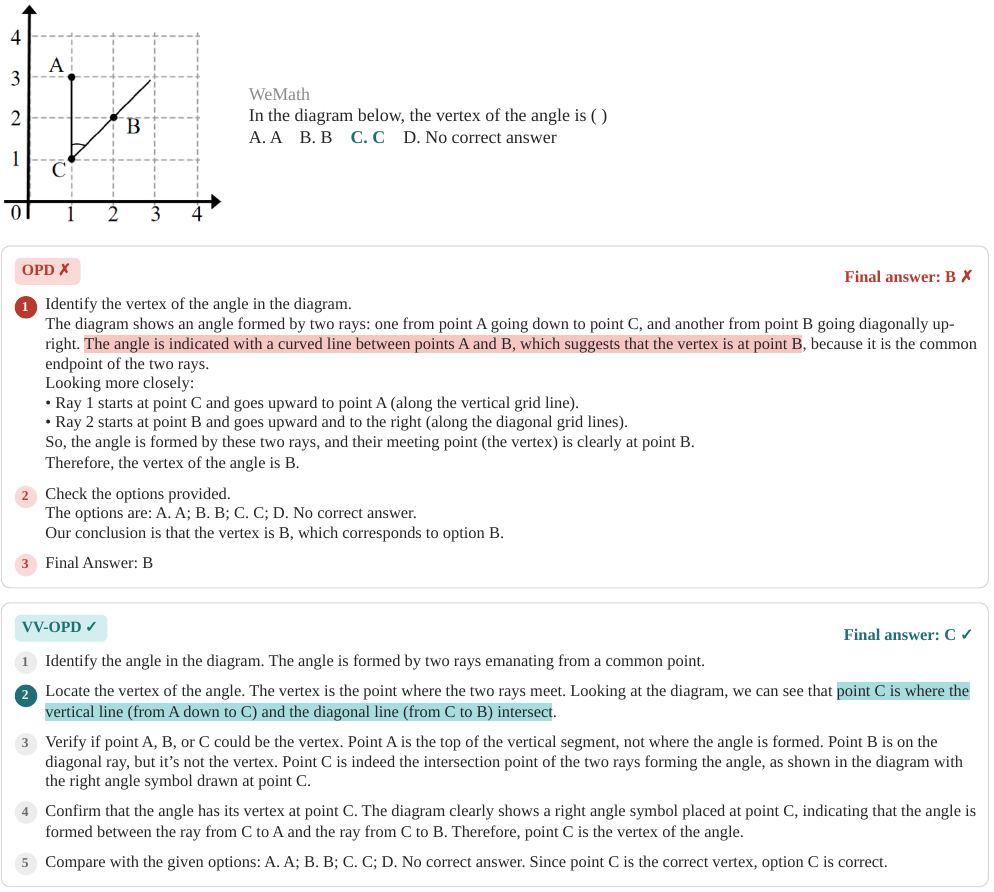}
  \caption{\textbf{Full reasoning traces for
  Fig.~\ref{fig:supp_qual_examples}(a)} (WeMath). Notation as in
  Fig.~\ref{fig:supp_trace_cylinder}.}
  \label{fig:supp_trace_angle}
\end{figure*}

\begin{figure*}[p]
  \centering
  \includegraphics[width=1.0\textwidth]{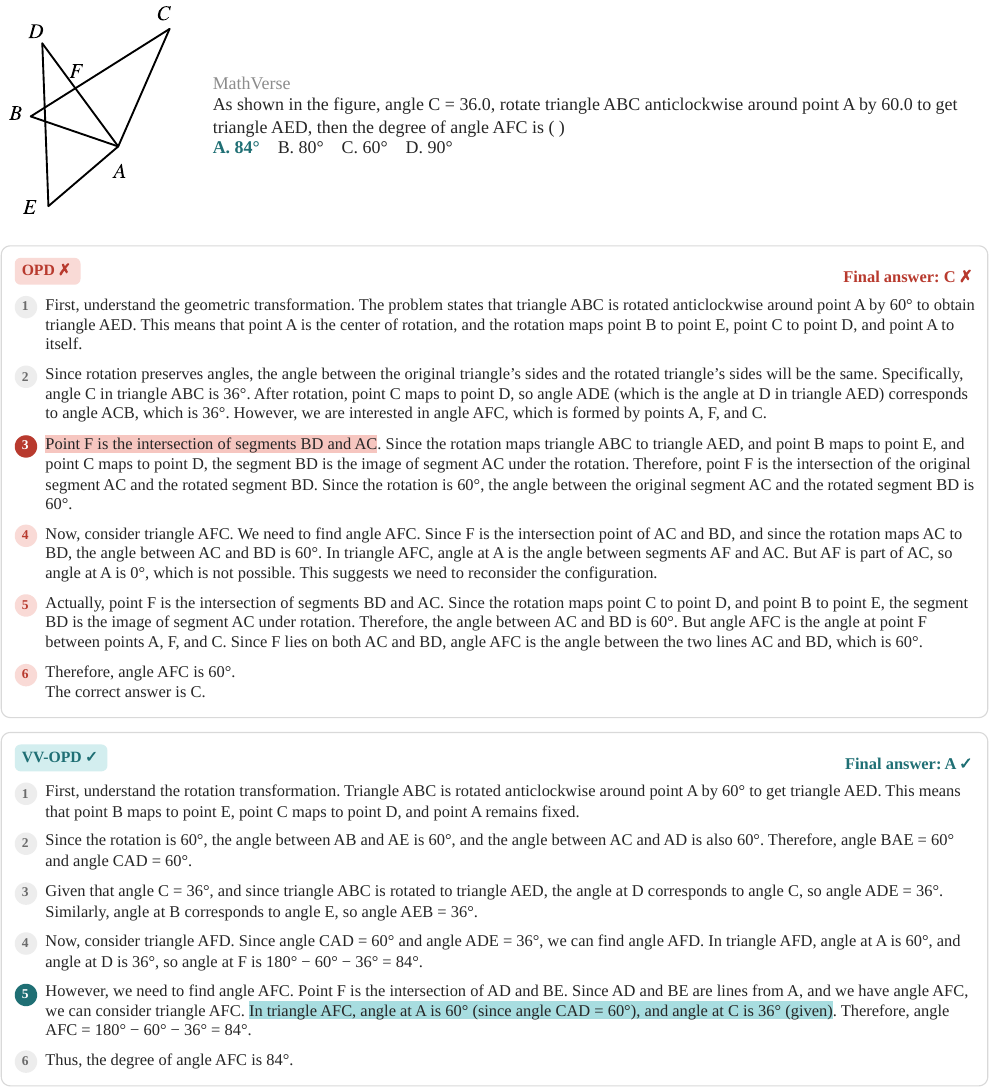}
  \caption{\textbf{Full reasoning traces for
  Fig.~\ref{fig:supp_qual_examples}(b)} (MathVerse). Notation as in
  Fig.~\ref{fig:supp_trace_cylinder}.}
  \label{fig:supp_trace_rotation}
\end{figure*}

\begin{figure*}[p]
  \centering
  \includegraphics[width=1.0\textwidth]{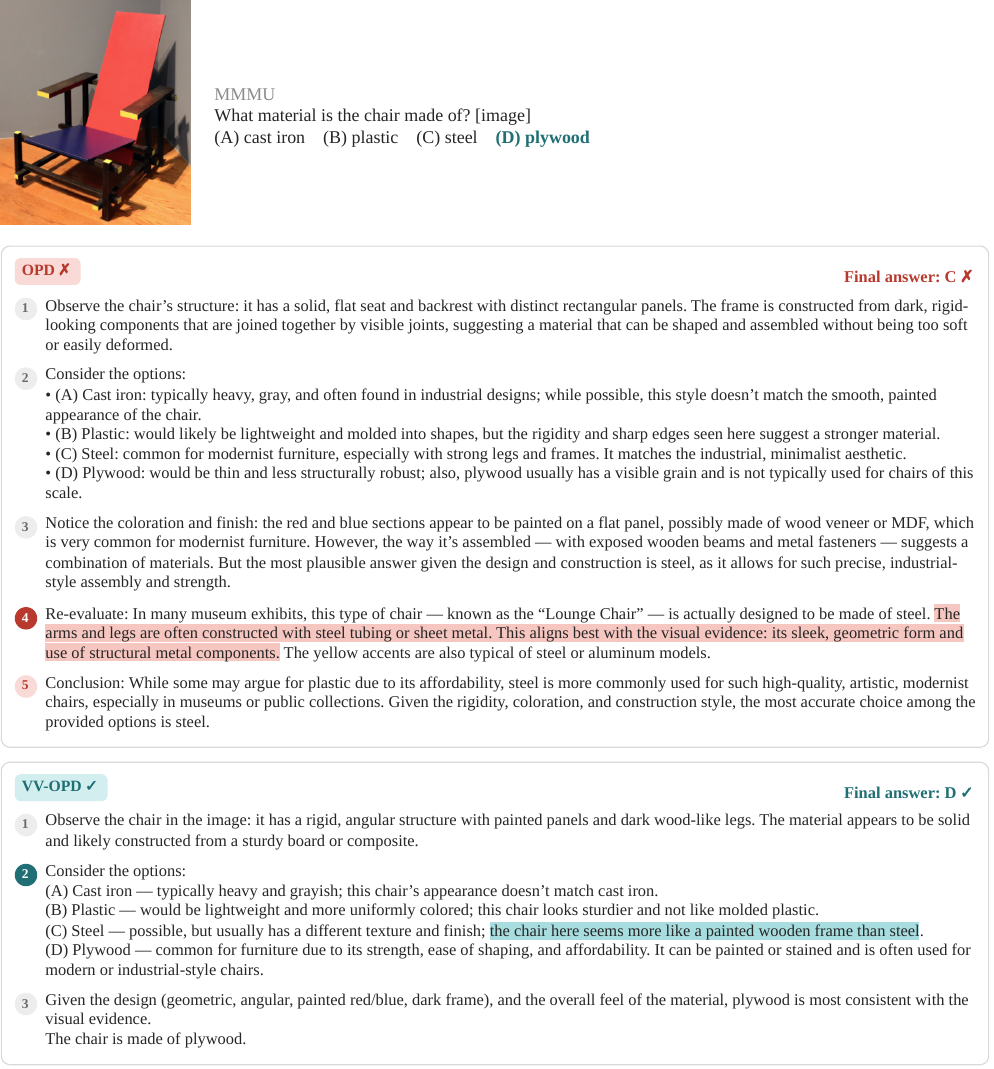}
  \caption{\textbf{Full reasoning traces for
  Fig.~\ref{fig:supp_qual_examples}(c)} (MMMU). Notation as in
  Fig.~\ref{fig:supp_trace_cylinder}.}
  \label{fig:supp_trace_chair}
\end{figure*}

\begin{figure*}[p]
  \centering
  \includegraphics[width=1.0\textwidth]{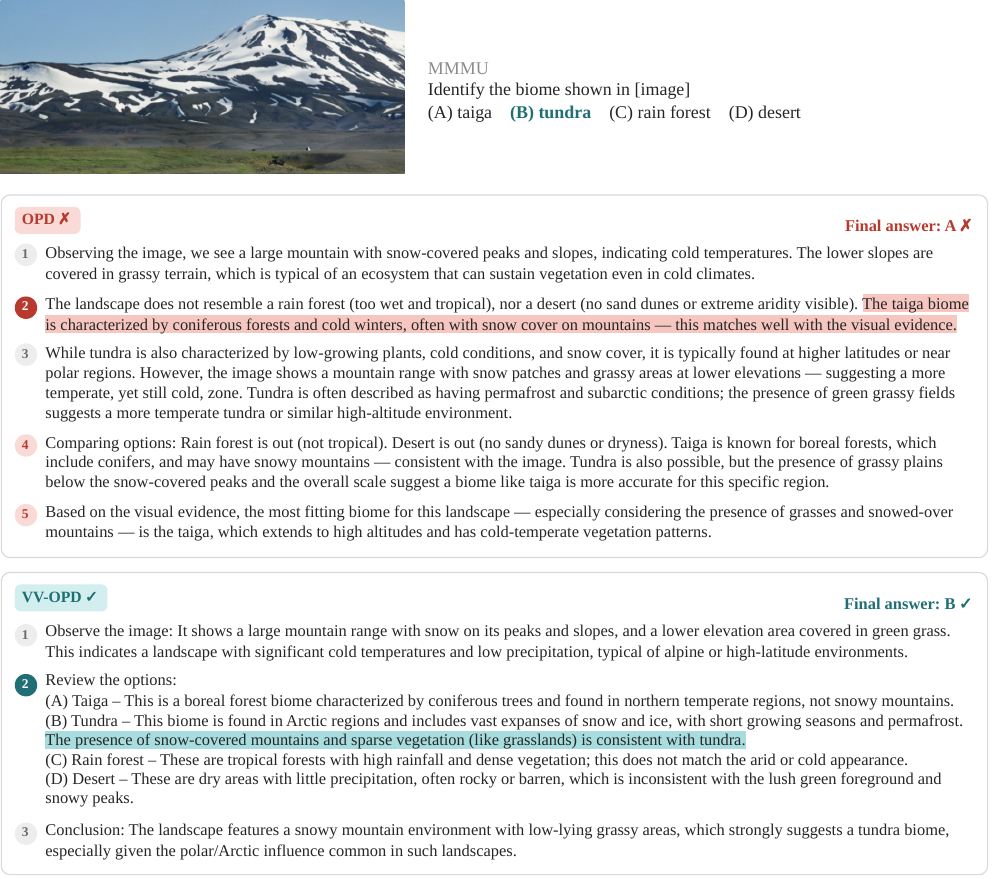}
  \caption{\textbf{Full reasoning traces for
  Fig.~\ref{fig:supp_qual_examples}(d)} (MMMU). Notation as in
  Fig.~\ref{fig:supp_trace_cylinder}.}
  \label{fig:supp_trace_biome}
\end{figure*}

\noindent\textbf{Full reasoning traces.}
Figs.~\ref{fig:supp_trace_cylinder}--\ref{fig:supp_trace_biome} list the
complete OPD and VV-OPD responses for the two main-paper examples
(Fig.~7) and the four examples above.
The steps are shown as generated, with formatting adjusted for
readability (e.g., typeset mathematical notation).
Dark badges mark the step containing the disputed visual claim,
and light red badges mark OPD steps that build on its error.
In Fig.~7(a), OPD reads the 5\,cm label as the diameter in step~2
and carries the resulting radius of 2.5\,cm into its volume
computation (Fig.~\ref{fig:supp_trace_cylinder}).

\end{document}

%% file: preamble.tex
\usepackage{natbib}
\setcitestyle{authoryear,square}
\usepackage[normalem]{ulem}
\usepackage{cancel}
\usepackage{verbatim}
\usepackage{float}
\usepackage{subcaption}
\usepackage{empheq}

\usepackage{booktabs}
\usepackage[table]{xcolor}
\usepackage{graphicx}
\usepackage{makecell}
\usepackage{multirow}

\newcommand{\verchain}{\mathrm{VER}_{\text{chain}}}
\newcommand{\verperc}{\mathrm{VER}_{\text{perc}}}

\definecolor{mathbg}{RGB}{226,226,250}
\definecolor{visbg}{RGB}{252,233,216}
\definecolor{avgbg}{RGB}{238,238,238}
\definecolor{gainbg}{RGB}{220,244,220}
\definecolor{gainfg}{RGB}{40,140,60}

\newcommand{\scoreup}[1]{%
  {\setlength{\fboxsep}{0.8pt}%
  \colorbox{gainbg}{%
    \textcolor{gainfg}{\scriptsize\bfseries$\uparrow$#1}}}}

\definecolor{vecrgainbg}{HTML}{E4F5EA}
\definecolor{vecrgainfg}{HTML}{217A55}

\usepackage[most]{tcolorbox}
\definecolor{promptteal}{HTML}{238C91}

\definecolor{vvteal}{HTML}{A8DDE0}     
\definecolor{mutedgrey}{HTML}{8C8C8C}

\usepackage{algorithm}
\usepackage{algpseudocode}

\algrenewcommand{\algorithmiccomment}[1]{\hfill{\color{gray}\(\triangleright\)~#1}}
\algrenewcommand\algorithmicindent{1.1em}

%% file: university_logos.tex
\newif\ifShowYaleLogo
\ShowYaleLogofalse

\newcommand{\YaleLogoHeight}{20pt}

\newcommand{\AdelaideLogoHeight}{20pt}
\newcommand{\CurtinLogoHeight}{18pt}
\newcommand{\SydneyLogoHeight}{20pt}

\newcommand{\LogoHeight}{20pt}

\newlength{\HeaderLogoBoxHeight}
\newcommand{\HeaderLogo}[2][]{%
  \raisebox{\dimexpr0.5\HeaderLogoBoxHeight-0.5\height\relax}%
    [\HeaderLogoBoxHeight][0pt]{%
      \includegraphics[#1]{#2}%
    }%
}

\renewcommand{\firstpagelogos}{%
  \makebox[\textwidth][s]{%

    \ifShowYaleLogo
      \HeaderLogo[
        height=\YaleLogoHeight,
        keepaspectratio
      ]{figs/logos/yale_logo.png}%
      \hfill
    \fi

    \HeaderLogo[
      height=\AdelaideLogoHeight,
      keepaspectratio
    ]{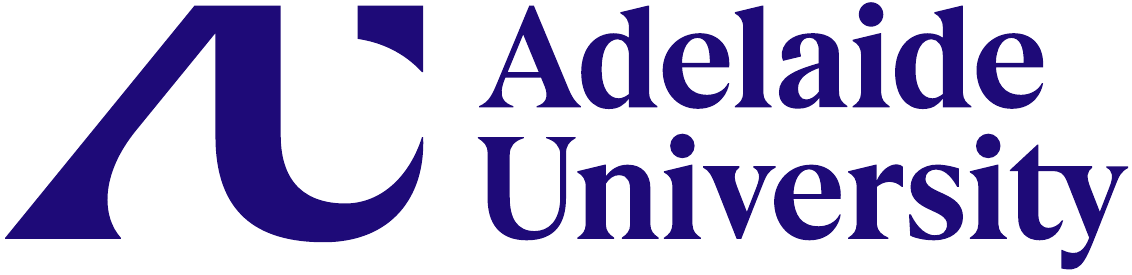}%

    \hfill

        \HeaderLogo[
      height=\SydneyLogoHeight,
      keepaspectratio
    ]{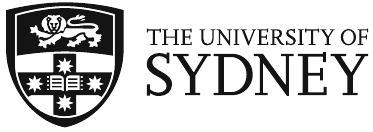}%
\hfill
    \HeaderLogo[
      height=\CurtinLogoHeight,
      keepaspectratio
    ]{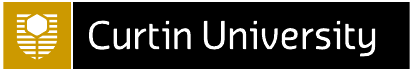}%

    \hfill

    \HeaderLogo[
      height=\LogoHeight,
      keepaspectratio
    ]{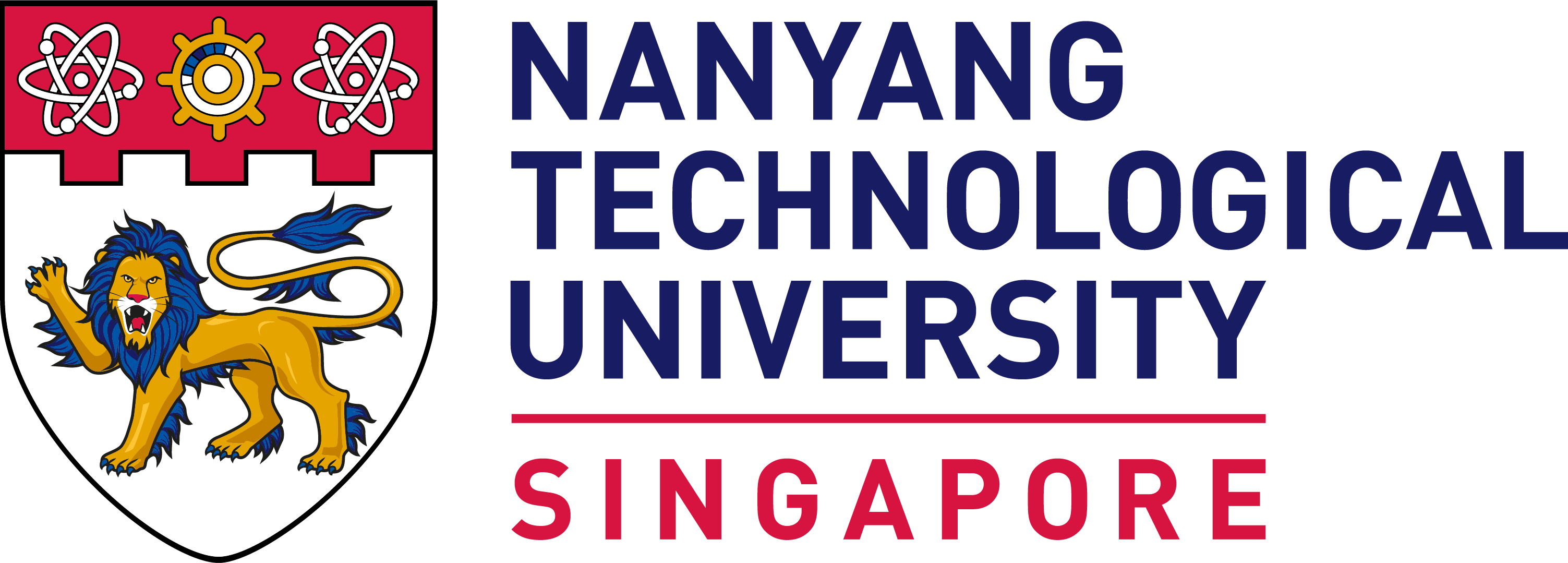}%

    \hfill

    \HeaderLogo[
      height=\LogoHeight,
      keepaspectratio
    ]{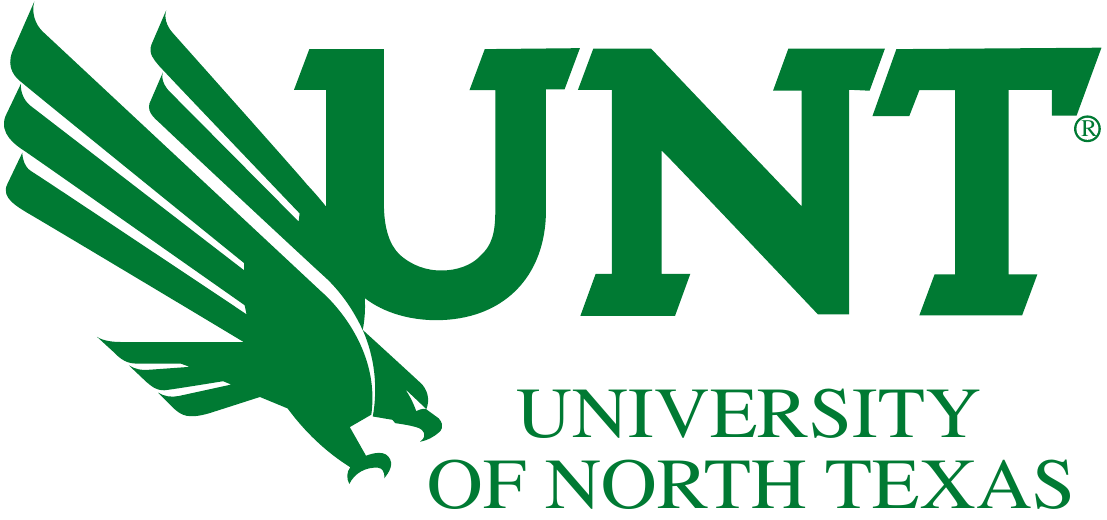}%

  }%
}